\documentclass[10pt, conference, letterpaper]{IEEEtran}
\IEEEoverridecommandlockouts

\usepackage{cite}

\ifCLASSINFOpdf
    \usepackage[pdftex]{graphicx}
    \usepackage{subfig}
\else
\fi

\usepackage{amsmath}
\usepackage{comment}
\usepackage{array}
\usepackage{multirow}
\usepackage{tabularx}
\usepackage{xspace}
\usepackage[capitalize]{cleveref}
\usepackage{siunitx}
\usepackage{amsmath}
\usepackage{mathtools, cuted}
\usepackage{float}
\usepackage{graphicx}
\newcommand{\name}{Firefly\xspace}
\newcommand{\indirecth}{Indirect-H\xspace}

\newcommand{\pso}{3D PSO\xspace}

\usepackage{color}
\newcommand{\todo}[1]{{\color{red}{\bf TODO:} #1}}
\newcommand{\tocheck}[1]{{\color{red}{} #1}}

\newcommand{\Ricardo}[1]{{\color{red}{\bf Notes from Ricardo:} #1}}

\begin{document}

\title{\name: Supporting Drone Localization With Visible Light Communication}

\author{
\IEEEauthorblockN{
Ricardo Ampudia Hernández \IEEEauthorrefmark{1},
Talia Xu\IEEEauthorrefmark{2},
Yanqiu Huang\IEEEauthorrefmark{3}
and
Marco A. Zúñiga Zamalloa\IEEEauthorrefmark{4}
}
\\
\IEEEauthorblockA{
\IEEEauthorrefmark{1}
\IEEEauthorrefmark{3}
Department of Computer Science,
University of Twente,
The Netherlands}
\IEEEauthorblockA{
\IEEEauthorrefmark{2}
\IEEEauthorrefmark{4}
Department of Computer Science,
TU Delft,
The Netherlands}
\thanks{The work was performed while the first author worked on his Master Thesis at TU Delft.}
}

% conference papers do not typically use \thanks and this command
% is locked out in conference mode. If really needed, such as for
% the acknowledgment of grants, issue a \IEEEoverridecommandlockouts
% after \documentclass

% for over three affiliations, or if they all won't fit within the width
% of the page, use this alternative format:
% 
%\author{\IEEEauthorblockN{Michael Shell\IEEEauthorrefmark{1},
%Homer Simpson\IEEEauthorrefmark{2},
%James Kirk\IEEEauthorrefmark{3}, 
%Montgomery Scott\IEEEauthorrefmark{3} and
%Eldon Tyrell\IEEEauthorrefmark{4}}
%\IEEEauthorblockA{\IEEEauthorrefmark{1}School of Electrical and Computer Engineering\\
%Georgia Institute of Technology,
%Atlanta, Georgia 30332--0250\\ Email: see http://www.michaelshell.org/contact.html}
%\IEEEauthorblockA{\IEEEauthorrefmark{2}Twentieth Century Fox, Springfield, USA\\
%Email: homer@thesimpsons.com}
%\IEEEauthorblockA{\IEEEauthorrefmark{3}Starfleet Academy, San Francisco, California 96678-2391\\
%Telephone: (800) 555--1212, Fax: (888) 555--1212}
%\IEEEauthorblockA{\IEEEauthorrefmark{4}Tyrell Inc., 123 Replicant Street, Los Angeles, California 90210--4321}}

% use for special paper notices
%\IEEEspecialpapernotice{(Invited Paper)}

% make the title area
\maketitle

% As a general rule, do not put math, special symbols or citations
% in the abstract
\begin{abstract}
Drones are not fully trusted yet. Their reliance on radios and cameras for navigation raises safety and privacy concerns. These systems can fail, causing accidents, or be misused for unauthorized recordings. Considering recent regulations allowing commercial drones to operate only at night, we propose a radically new approach where drones obtain navigation information from artificial lighting. In our system, standard light bulbs modulate their intensity to send beacons and drones decode this information with a simple photodiode. This optical information is combined with the inertial and altitude sensors in the drones to provide localization without the need for radios, GPS or cameras. Our framework is the first to provide 3D drone localization with light and we evaluate it with a testbed consisting of four light beacons and a mini-drone. We show that, our approach allows to locate the drone within a few decimeters of the actual position and compared to state-of-the-art positioning methods, reduces the localization error by 42\%.
\end{abstract}

% no keywords

% For peer review papers, you can put extra information on the cover
% page as needed:
% \ifCLASSOPTIONpeerreview
% \begin{center} \bfseries EDICS Category: 3-BBND \end{center}
% \fi
%
% For peerreview papers, this IEEEtran command inserts a page break and
% creates the second title. It will be ignored for other modes.
\IEEEpeerreviewmaketitle

\section{Introduction} 

Delivery with autonomous drones represents a fascinating future not only for commercial products but also for medical and food supply~\cite{Vital, Matternet}. However, as illustrated in a recent report by McKinsey~\cite{Duvall2021}, making this future a reality faces many challenges in range, safety and infrastructure support. People still feel nervous, scared or even angry under the presence of drones and only 11\% of users think that drones should be allowed near homes~\cite{hitlin_2020}. Due to these reservations, most of the world still has restrictive commercial regulations, from outright bans to limited experimental licenses~\cite{jones_2017}. 

A broad commercial adoption of drones is hindered because people have concerns about safety and privacy. Drone operation depends on three key components: GPS~\cite{Patrik2019}, RF wireless links~\cite{chao_cao_chen_2010}, and cameras~\cite{Lu2018}. But what if one of these components fails or its use is prohibited in certain areas? For example, GPS is known to face limitations indoors and in urban canyons~\cite{NASA2020}; RF signals face ever-increasing spectrum saturation and are prone to interference; and cameras raise privacy concerns (making them undesirable in various areas~\cite{rice_2019}) and consume orders of magnitude more power than simpler photosensors (impacting the lifetime of mini drones).

\textbf{Opportunity.} A pervasive and dependable infrastructure is crucial for reliable and safe drone operation, yet it has been largely overlooked~\cite{Duvall2021}. Compared to autonomous vehicles, which build upon a comprehensive road network with sensors, cameras, traffic lights and road signs, unmanned drones do not have any such support. To minimize the risk to citizens, recent regulations are allowing drones to fly only \textbf{at night}~\cite{Transportation2021}. This regulatory framework, effective in April 2021, opens the opportunity of transforming the vast presence of lighting in our cities into a pervasive infrastructure for drone navigation.

\textbf{Vision.} Similar to the way old lighthouses provided navigational aid to maritime pilots, \name aims at transforming standard light bulbs --such as those present in our roads, streets and buildings-- into a modern version of those lighthouses. Light bulbs will play the role of “air traffic control towers”, modulating their intensities to provide navigation services to drones. 
To achieve this goal, Firefly exploits recent advances in visible light communication (VLC)~\cite{tsonev_videv_haas_2013}, an emerging technology that transmits information using any type of LED. The core idea behind VLC is similar to communicating with somebody by turning a flashlight on and off. The modulation, however, is done at such high speeds that humans only see a normal light on (no flickering), while drones will be able to decode digital information using a simple photodiode.

\textbf{Contributions.} The use of visible light communication for drones is largely unexplored. A few theoretical studies have proposed the use of drones to provide temporary Internet coverage to people (i.e., mobile lights acting as access points~\cite{yang_chen_guo_feng_saad_2019}), and a more recent study assesses the link between a drone’s camera and a ground light~\cite{Chhaglani2020}. These studies focus solely on analyzing the communication link with visible light but do not provide any type of localization service to the drone. On the other hand, various methods have been proposed to use visible light for indoor positioning, but our study is the first to show accurate 3D positioning in scenarios with six degrees of freedom. Overall, \name provides two main contributions.

%\vspace{1mm}
\textit{Contribution 1: Analytical Framework [Sections \ref{sec:Analysis} \& \ref{sec:method}].} Considering that the state-of-the-art already provides notable contributions on visible light positioning~\cite{Almadani-2018-VLC-industrial-environment, Do-2016-in-depth-survey-vlp, zhuang2018survey-vlp-leds, afzalan2019-survey-real-prototypes}, we first perform a thorough analysis of the existing localization primitives and show that the approach decomposing the 3D problem into a 2D+H problem (where H stands for height) is the best alternative~\cite{plets2019efficient, Almadani-2018-VLC-industrial-environment}. After that, we identify the limitations of the 2D+H approach on drones (due to the high number of degrees of freedom) and propose a framework that combines visible light information, together with the inertial and altitude sensors in drones, to attain accurate 3D localization.   
A key property of our framework is the low complexity of the HW (transmitters and receiver) and SW (localization method). 
%An important property of our framework is the simplicity of the transmitters (light beacons), receiver (sensor on drone) and the low complexity of the localization method. 

%\vspace{1mm}
\textit{Contribution 2: Platforms \& Evaluation [Section~\ref{sec:Experimental_Validation}].} We evaluate our framework using empirical data. We build a testbed consisting of four light beacons and add a PCB with a single photodiode to a mini-drone. The testbed works in the dark as well as in scenarios exposed to ambient light. %\tocheck{Our main result shows that the localization accuracy of drones can be reduced from more than a \SI{0.4}{\m} (using the default 2D+H approach) to a few decimeters using \name.}
Our main result shows that the localization of drones can be achieved with an accuracy of a few decimeters and improved by 42\% compared to available methods in the state-of-the-art.
\begin{figure*}[h]
	\begin{minipage}[H]{0.23\linewidth}
		\centering
		{\includegraphics[width=\columnwidth, height = \columnwidth]{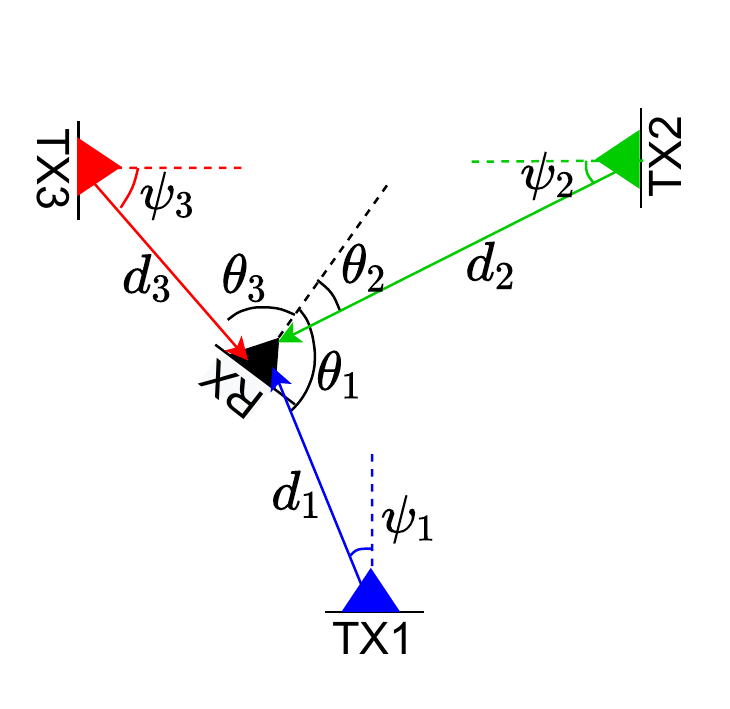}}
		\caption{Basic localization with LEDs.}
		\label{fig:LED-tri}
	\end{minipage}
	\hspace{0.1cm}
	\begin{minipage}[H]{0.23\linewidth}
		\centering
		{\includegraphics[width=\columnwidth, height = \columnwidth]{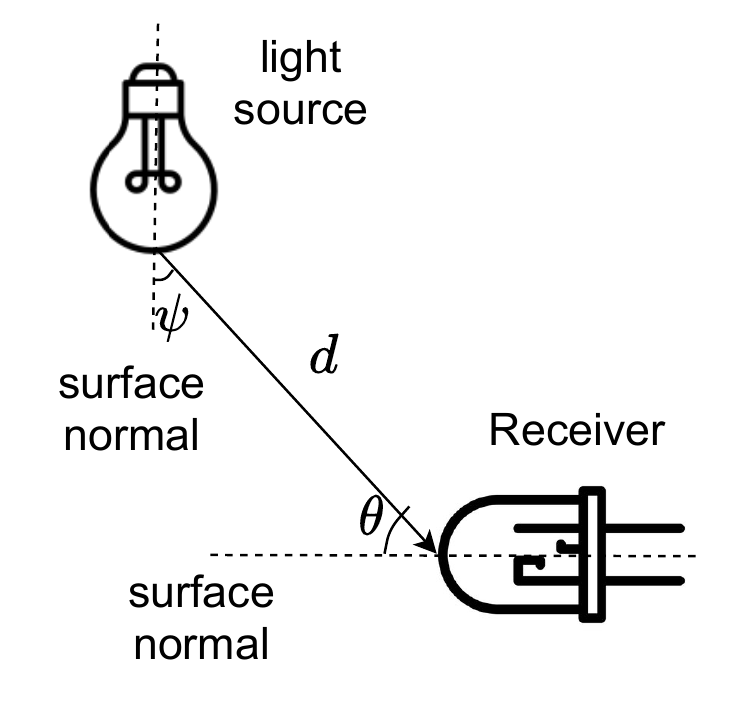}}
		\caption{LED propagation properties.}
		\label{fig:LED-lambertian}
	\end{minipage}
	\hspace{0.1cm}
	\begin{minipage}[H]{0.23\linewidth}
		\centering
		{\includegraphics[width=\columnwidth, height = \columnwidth]{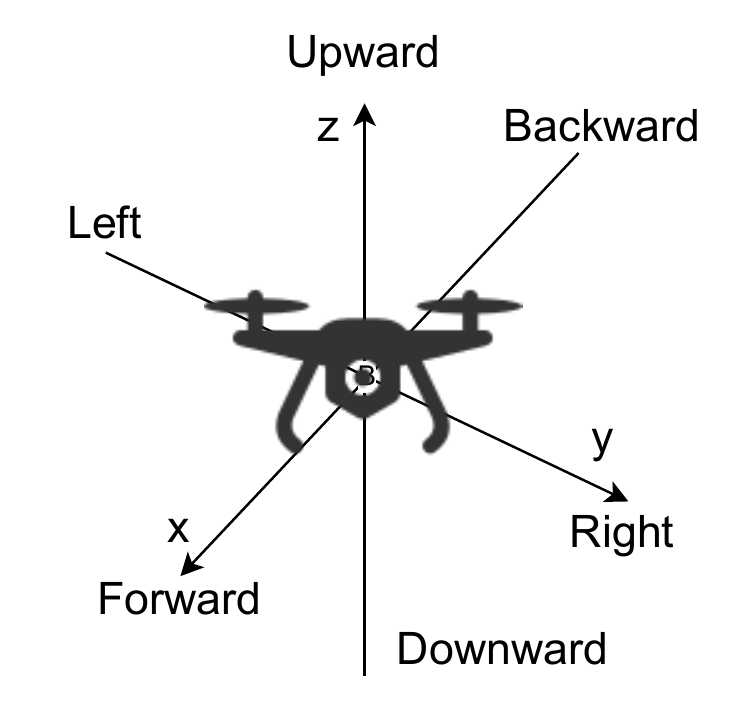}}
		\caption{Transitional\\movement.}
		\label{fig:Drone-transitional}
	\end{minipage}
	\hspace{0.1cm}
	\begin{minipage}[H]{0.23\linewidth}
		\centering
		{\includegraphics[width=\columnwidth, height = \columnwidth]{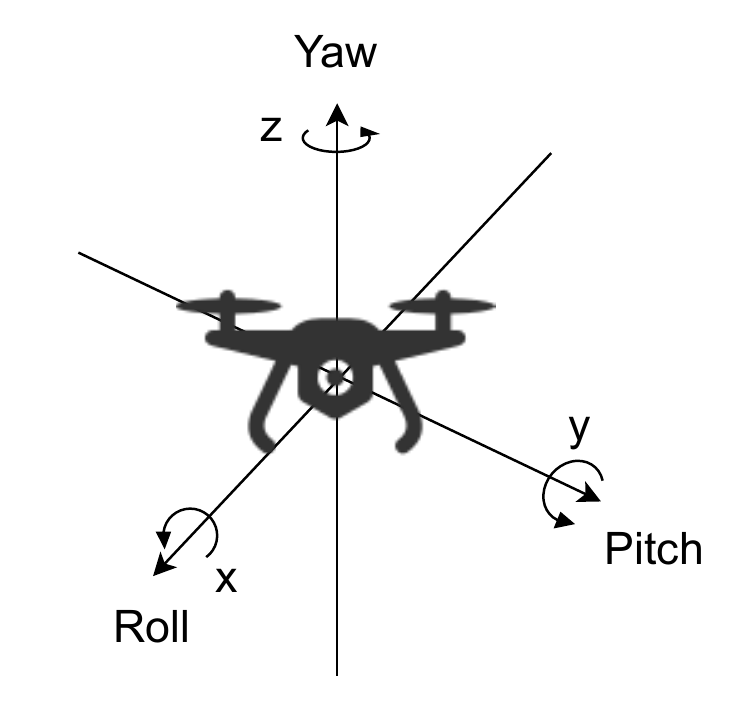}}
		\caption{Rotational\\movement.}
		\label{fig:Drone-rotational}
	\end{minipage}
	\hspace{0.1cm}
\end{figure*}

\section{Background} %Talia
\label{sec:background}

%Lambertian model and its use for VLP ... \\
%Barometer? Accelerometer? Inertial navigation? Sensor fusion?\\

In this section, we present the background information on the Lambertian patterns of LED lights, as well as the basic localization principles behind this work. We introduce everything in the 2D space for clarity, however, these concepts are also applicable to the 3D space.

Two types of localization techniques are often used in visible light positioning (VLP): ones that make use of the received signal strength (RSS), and ones  that  make  use  of  the  angle  of  arrival  (AOA).  In  each method,  the  received  power  or  angle  information  acquired from  different  light  sources  is  used  to  estimate  the position of a receiver. In this work, we focus on RSS methods for reasons detailed in \cref{sec:Analysis}.

In VLP with RSS, LED lights are used as anchor points with known locations, as shown in Fig.~\ref{fig:LED-tri}. Each LED light (represented by TX1 to TX3) broadcasts a beacon and the receiver measures the RSS of each signal. The receiver uses the received power to estimate its distance to the different light sources and obtains its location through a trilateration method. 
Starting at the transmitter, the propagation of light follows a Lambertian pattern $R_t(\psi)$, which is determined by the equation below and also depicted in Fig.~\ref{fig:LED-lambertian}.

\begin{equation}
	R_t(\psi) = \frac{m + 1}{2\pi} \cos^m(\psi),
	\label{eq:lambertian}
\end{equation}
where $\psi$ represents the direction relative to the surface normal of the transmitter, %as shown in Fig.~\ref{fig:LED-lambertian}, %$R_t(\psi)$ represents the propagation pattern of the LED light 
and $m$ represents the Lambertian order. The Lambertian order $m$ determines the width of the beam from the transmitter. A small value of $m$ leads to a wider beam, while a larger $m$ leads to a narrow beam, similar to a spotlight.

The Lambertian pattern defines the optical wireless channel between the transmitter and the receiver, and it is described in the following equation:
\begin{equation}
	H(0) = \begin{cases}
		R_t(\psi) \cdot A_\text{r} \cdot \frac{\cos \theta}{d^2} & \text{for} \hspace{0.25cm} \theta \in [0, \Theta_c],\\
		0 & \text{for} \hspace{0.25cm} \theta > \Theta_c,
	\end{cases}
	\label{eq_chloss}
\end{equation}
where $d$ is the distance between the transmitter and the receiver; $\theta$ denotes the incidence angle at the receiver; $A_{r}$ is the effective sensing area of the receiver; and $\Theta_c$ represents the field-of-view (FoV) of the receiver, beyond which the receiver is unable to detect the incoming signal. Combining the propagation pattern of the transmitter and the effect of the channel, the received power $P_r$ at the PD can be written as:
\begin{equation}
	P_r = P_t \cdot H(0)\cdot g_r(\theta) + N ,
	\label{eq:rxowerp_w_nf}
\end{equation}
where $P_t$ is the transmitted power; $g_r(\theta)$ is the optical gain of the PD~\cite{Yin2015}; $N$ represents the sum of the shot and thermal noise of the receiver, as well as the ambient noise.
%and $P_t$ and $P_r$ are the transmitted and received power, respectively. 

In VLP for drones, the mobile receiver has 6 degrees of freedom (DoF), as shown in Fig.~\ref{fig:Drone-transitional} and Fig.~\ref{fig:Drone-rotational}. Compared to the setups of state-of-the-art (SoA) studies, which typically consider static receivers, the movements of drones introduce two important dynamics. First, the distance $d$ to a transmitter changes, affecting the irradiance angle $\psi$. Second, any transitional movement requires the drone to tilt around its axes, affecting the incidence angle $\theta$. For example, in order to move forward and backward, the drone has to tilt in the pitch axis; and the tilting angle influences the speed of the movement.

The above two effects can have a significant impact on the RSS ($P_r$) and the location estimation. For example, considering only a \SI{5}{\degree} deviation in the angles, between 20$^\circ$ and 15$^\circ$, the distance $d$ will be overestimated by more 10\%, around 1.5\% due to the misalignment at the receiver (effect of $\cos(\theta)$ in \cref{eq_chloss}) and 8.5\% due to the misalignment at the transmitter (effect of $\cos^m(\psi)$ in \cref{eq:lambertian} assuming a Lambertian order $m=6$). Thus, compared to a static scenario where the receiver is not moving and the angles are known and fixed, the constant movement and tilting in drones make localization more challenging. \name provides a simple method based on the equations presented in this section and solely based on parameters available in the data sheets of the transmitters and receivers (i.e., $m$, $A_\text{RX}$ and $g_r(\theta)$).
\section{Analysis of the State of the Art} %Ricardo
\label{sec:Analysis}

In this section, we introduce the available VLP techniques in three categories: AOA-based 3D, RSS-based 3D and RSS-based 2D+H. %We will refer to the last category as \indirecth and 
We analyze the feasibility of each category for drone localization using the following criteria:

\begin{itemize}
    \item The complexity of the transmitter design, which is critical if we are to transform our lighting system with minimal changes.
    \item The complexity of the receiver design considering the limited weight capacity and power budget of drones.
    \item  The complexity of the algorithm, which affects the processing, memory and power requirements in the constrained environment of the drone.
    \item Whether tilting is considered, which is important for mobile scenarios as discussed in \cref{sec:background}.
\end{itemize}
\vspace{1mm}

\subsection{3D VLP with AOA}
\label{sec:Analysis-AOA}

In previous studies, AOA-based methods have shown to be more accurate than RSS-based methods for 3D VLP in simulation and experimental evaluations in static conditions, achieving up to sub-decimeter levels of accuracy in certain scenarios~\cite{csahin2015hybrid, zhu2018-vlp-adoa-tilting, yang2014three}. However, AOA-based methods have considerably higher complexities in terms of the required infrastructure support, the underlying mathematical framework and the computational complexity. For example, in the design by Yang et al.~\cite{yang2014three}, and the design by Xie et al.~\cite{xie2016lips}, multiple PDs are arranged at different angles in polygonal structures to allow the use of both AOA and RSS information to uniquely determine the position of a receiver. The requirement of using multiple PDs placed at different angles increases the profile of the receiver significantly and may not be feasible on small drones. In the design by Zhu et al.~\cite{zhu2018-vlp-adoa-tilting}, very accurate results are reported based on the angle difference of arrival (ADOA) information. However, in addition to using a PD array, a complex framework is also required, which results in a high computation time even on a dual-core \SI{2.4}{GHz} processor. In the design by Sahin et al.~\cite{csahin2015hybrid}, the complexity of design is transferred from the receiver to the transmitter by proposing the use of Visible Light Access Points (VAPs). VAPs consist of multiple tilted LEDs, which are used to provide AOA information to a single receiver sensor on the drone. However, this custom design makes the transmitter difficult to scale. 

In summary, AOA-based methods can achieve a high accuracy but require either complex receiver designs and algorithms, which are infeasible for small drones; or elaborated and costly transmitters, which do not scale well to a large lighting infrastructure.  %dedicated infrastructure, which increases the number of transmitters and receivers, and require complex hardware and algorithmic frameworks. This results in high processing and energy stress for 3D VLP for drones.

\subsection{3D VLP with RSS}

Compared to AOA-based methods, RSS-based methods require considerably less infrastructure but impose some stringent constraints on the evaluation setup. %\todo{mobility (remove ``mobility" and use ``the evaluation setup" if AoA receivers are static).} \Ricardo{"Experimental setup" instead?}. 
For example, in the study by Li et al.~\cite{li2014epsilon} and the study by Zhuang et al.~\cite{zhuang2019-vlp-noise-mitigation}, the receiver and the transmitter are assumed to be parallel to simplify the problem. Even though the design implementations have lower complexities, the parallel assumption is unlikely to remain valid when drones are flying. In the study by Cai et al.~\cite{cai2017-vlp-pso}, a 3D RSS-based VLP method is demonstrated with an average error of a few centimeters. However, a computationally intensive particle swarm optimization (PSO) algorithm is adopted to solve for the position. In addition, the experimental scenarios consider that receivers remain static and parallel to the transmitters. In the study of Carreño et al.~\cite{carreno20-metaheuristic}, the computationally-heavy Genetic Algorithm (GA) is also considered to solve the 3D VLP problem and reports comparable accuracy results.

In summary, RSS-based methods have potential for 3D VLP for drones, as they require less hardware complexities compared to AOA-based methods. However, the main shortcomings are their computationally intensive optimization algorithms and the fact that the methods are accurate only with \textit{static} receivers maintaining a \textit{parallel orientation} with the transmitters. 

%are still required. In addition, the experimental evaluations for RSS-based methods have been very limited with only static receivers.

%\subsection{\indirecth VLP with RSS}
\subsection{2D+H VLP with RSS (\indirecth)}
\label{sec:SoA-2dIh}

A promising approach in using RSS-based methods for 3D VLP is to decompose the problem space into \textit{2D+H}, where the height (H) and the 2D position of the receiver are independently estimated. For example, in the studies by Plets et al.~\cite{plets2019efficient} and by Alamadani et al.~\cite{ almadani2019novel}, a list of different height values are sequentially assumed, and for each height, the 2D position is solved via trilateration. After the trilateration process is completed for all different heights, the algorithm then determines which height is the most probable to be correct by minimizing the error of a cost function that identifies which candidate solution is the most likely. Both studies are able to achieve 3D VLP with modest infrastructure and algorithmic complexities, and report favorable results for \textit{static} receivers that are parallel to the transmitters. In the study of \cite{almadani2019novel}, however, when tilting is introduced, the positioning error of their method increases more than 15 times when the tilting angle is 3$^\circ$ and more than 30 times when it is 5$^\circ$. In addition, the experiments are performed in a controlled environment using a high-end PD, which do not reflect the normal working conditions of drones.
%Using a \textit{2D+Indirect H} approach with RSS information leads to simple receivers, transmitters, and methods. The main shortcoming is that the systems has only been tested in static conditions and they only provide accurate results when the receiver is parallel to the transmitter.
A key problem of 2D+H methods is that they use an \textit{indirect} approach to estimate height. Height is obtained through RSS measurements. This approach works well with parallel and static receivers, but mobility and tilting affect their performance. We use these studies as the baseline of comparison for \name, and for the remainder of the paper we will refer to 2D+H Indirect methods as \indirecth.\\

\begin{figure}[!t]
    \centering
    \includegraphics[width=0.8\columnwidth]{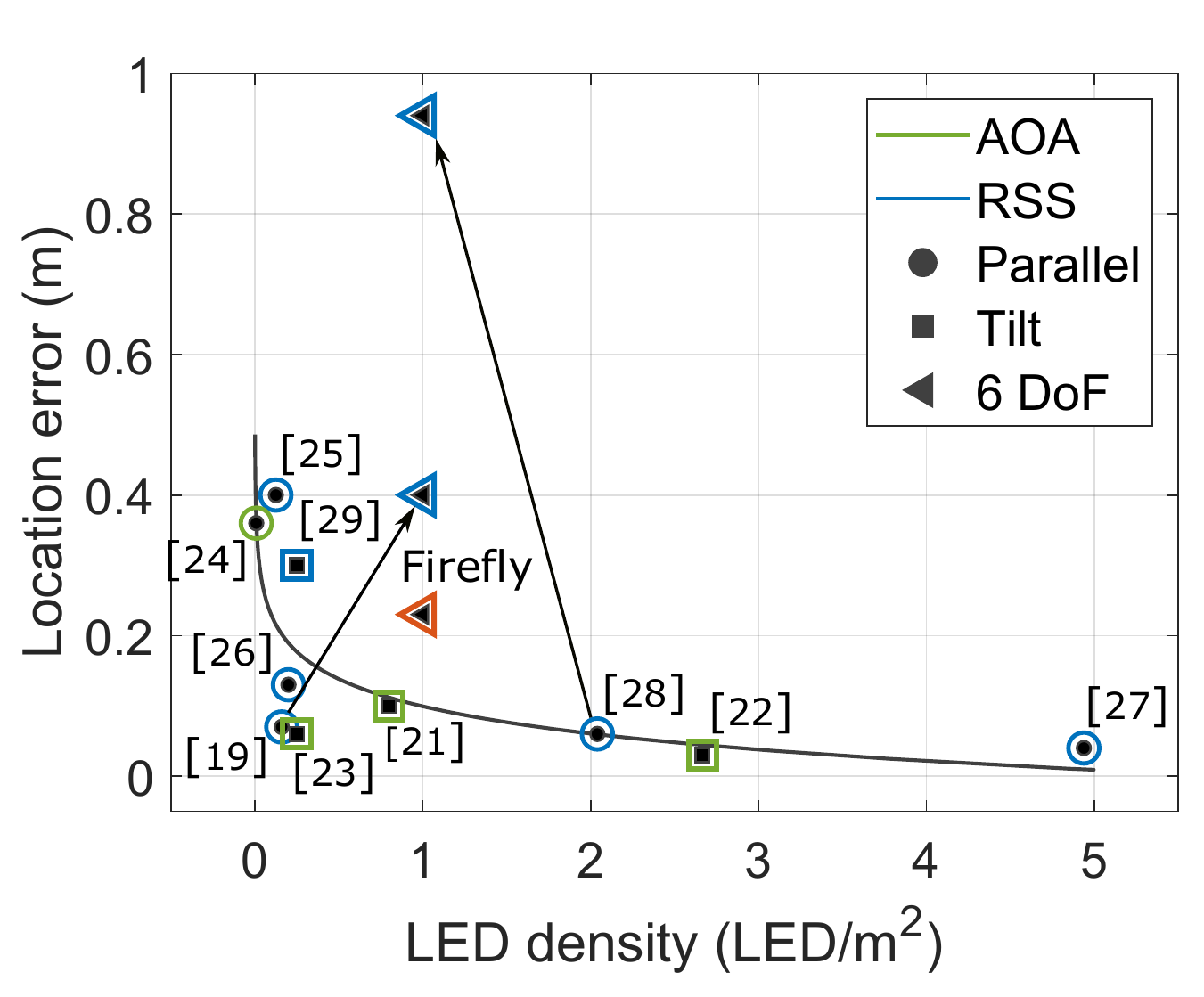} %blue square
    \caption{Accuracy vs LED density for different SoA studies. The curve represents the average trend for the static scenarios (without including \name and the two end points of the arrows evaluated in our drone testbed).}
    \label{fig:soa_graph}
\end{figure}

To conclude this section, we show the performance of the discussed SoA methods in terms of accuracy (reported positioning error) and the LED density in \cref{fig:soa_graph}. The density of anchor points is a major indicator of how well a VLP system will perform and provides a basis for comparison \cite{afzalan2019-survey-real-prototypes}.  %Some achieve higher accuracy at the expense of more LEDs per unit area. 
The methods are categorized by type (AOA or RSS) and testing conditions (if the receivers are \textit{parallel} to the transmitters in their evaluations or if they consider \textit{tilted} setups).
All these studies, including the ones evaluating tilting, consider only \textit{static} receivers in their experimental evaluations. In order to have a high localization accuracy for drones while keeping the hardware complexity low, we propose Firefly, which will be discussed in \cref{sec:method}. In \cref{ssec:experimental_setup} we compare \name against two other SoA approaches in a significantly more challenging \textit{mobile} scenario. In \cref{fig:soa_graph}, we can observe that when we test those two SoA methods with 6 DoF drones (arrows), their performance decrease dramatically.  %The blue triangle depicts the performance of \name, attained with low complexity of the design in a significantly more challenging \textit{mobile} scenario.

%%%%%%%%%%%% Previous version
\begin{comment}
To conclude, we show the performance of the discussed SOA methods in terms of accuracy (reported positioning error) and the \tocheck{LED density (a major indicator of how well a VLP systems can perform) in \cref{fig:soa_graph}. Some achieve higher accuracy at the expense of more LEDs per unit area}. All these studies, including the ones evaluating tilting, consider only \textit{static} receivers in their experimental evaluations. In order to have a high localization accuracy for mobile drones while keeping the hardware complexity low, we propose Firefly, which will be discussed in \cref{sec:method}. The blue triangle depicts the performance of \name, attained with low complexity of the design in a significantly more challenging \textit{mobile} scenario.
\end{comment}

\section{Proposed method}%Ricardo
\label{sec:method}

%    - Firefly aims to keep the harwdare compelxity low - RSS
%    - In SoA, Indirect H methods are considered to obtain the position of the drone. 
%    - We explain SoA
%    This methods relies in one important assumptions:
%        - Parallel
%    - However, looking into equation 3, we see that angle of the receiver plays and important role in the estimation.
%    - In the case of drones tilting is present
    
%    Our method consists of three main steps to address the incidence and irradiance angle.
%        - Height estimation
%        - Irradiance angle 
%        - Incidence angle 

In Firefly, our aim is to keep the hardware complexity as low as possible. Therefore, we consider an infrastructure where each LED transmitter has a single off-the-shelf light with no tilted or custom-made structure. This maintains the simplicity of the transmitters, making them easily scalable. Similarly, on the receiver side, our design requires each drone to be equipped with only a single PD for VLP. %As described in Sec.~\ref{sec:background}, we assume that the models and specifications of the LED transmitters and PD receiver are known. Therefore, with the RSS information at the PD, the distances between the transmitters and the PD on a drone can be calculated directly from Eq.~\ref{eq:rxowerp_w_nf}, if the incidence angle $\theta$ and irradiance angle $\phi$ are also known.

%In our system, we propose the use of a \textit{2D+Direct H} method with RSS for 3D VLP for drones. In this approach, we obtain direct and reliable height measurements by using inertial and barometric sensors, which are already present in most off-the-shelf drones. 

Based on the analysis of \cref{sec:Analysis}, we consider the \indirecth methods from~\cite{almadani2019novel} and~\cite{plets2019efficient} as the initial building block because, even though they assume that the transmitter and receiver have to be parallel to each other to obtain an accurate result, they have low hardware and algorithmic complexities. %These methods rely on the important premise that the transmitter and the receiver are parallel to each other, which is frequently not the case when a drone is flying. 
%These methods rely on two important assumptions (explained hereafter) that do not hold when tilting is present, which constantly happen when a drone is flying.
We will first describe the %\indirecth method 
common framework, assumptions and limitations of RSS methods, in particular the \indirecth approach. After that, we explain our method and the modifications to the common framework that enable accurate VLP in a mobile scenario. \\

\subsection{Common framework of RSS methods}
\label{ss:common_framework}
\begin{figure}[!t]
    \centering
    \includegraphics[width=0.9\columnwidth]{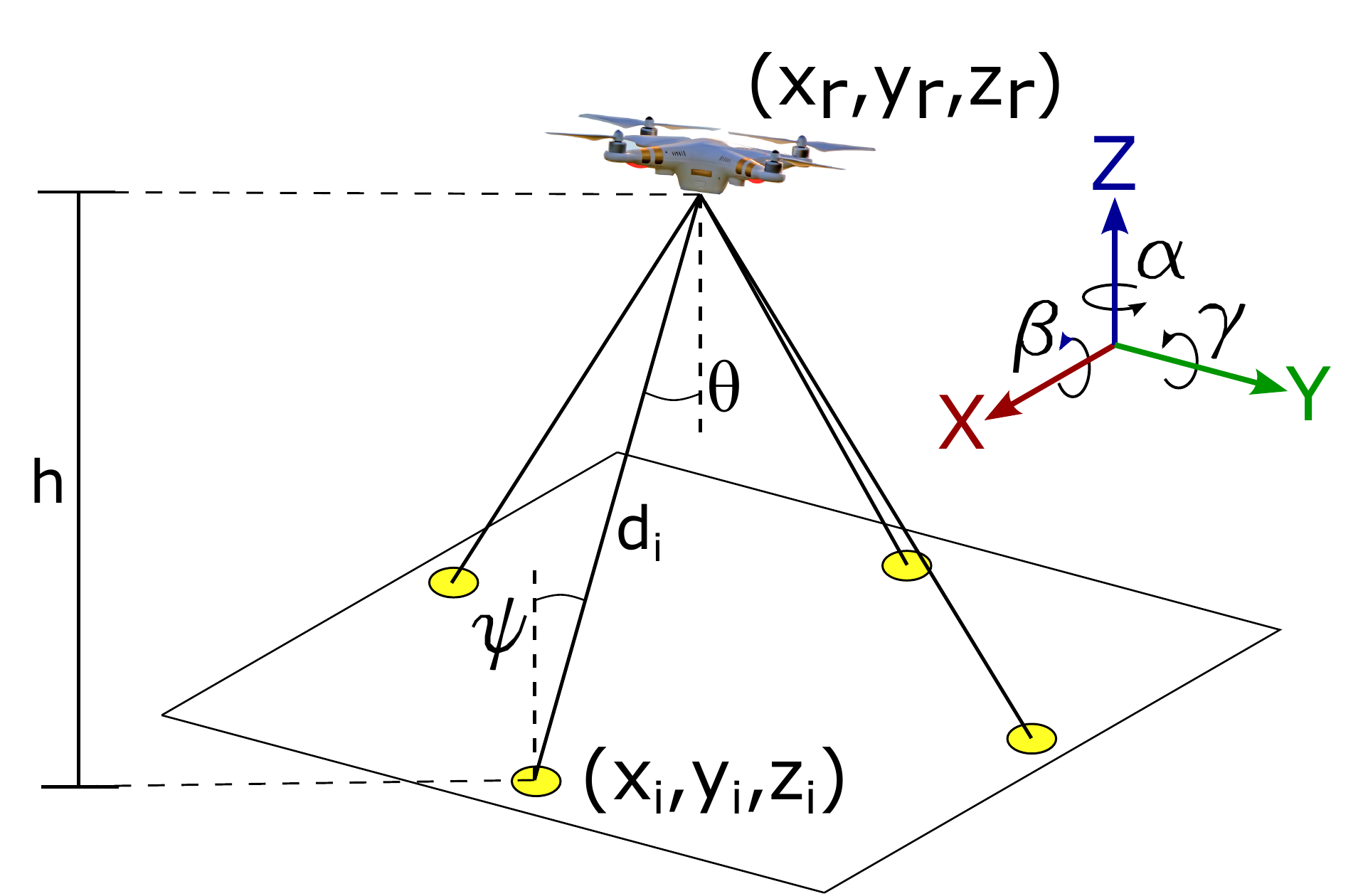}
    \caption{System reference and parameters. Without loss of generality, we place the LEDs (transmitters) on the ground.}
    \label{fig:drone_ref}
\end{figure}

As described in Sec.~\ref{sec:background}, we assume that the parameters from the data sheets of the LED transmitters and PD receiver are known. Therefore, with the RSS information at the PD, the distances between the transmitters and the PD on a drone can be calculated directly from Eq.~\ref{eq:rxowerp_w_nf}, if the incidence angle $\theta$ and irradiance angle $\phi$ are also known.
\begin{comment}
A common assumption in the SoA studies, including \indirecth methods, is that the receiver remains parallel ($\parallel$) to all transmitters, since this greatly simplifies the problem. With this assumption, we obtain the following equation:
\end{comment}
A common assumption in the SoA studies, including \indirecth methods, is
\begin{itemize}
    \item Assumption 1 (A1): The transmitters (TX) and receivers (RX) are parallel ($\parallel$).
\end{itemize}
This assumption greatly simplifies the problem and leads to the following equation:

\begin{equation}
    cos(\psi) = cos(\theta) = \frac{|z_r - z_i|}{d_{i, \parallel}} = \frac{h}{d_{i, \parallel}},
    \label{eqn:parallel_assumption}
\end{equation}
where $z_r$ denotes the height of the receiver; $z_i$ represents the height of the \textit{i}-th transmitter; and $d_{i,\parallel}$ is the distance between the receiver and the \textit{i}-th transmitter. In \cref{fig:drone_ref}, we show the reference and parameters of the system for clarity. By substituting \cref{eqn:parallel_assumption} into \cref{eq:rxowerp_w_nf}, the distance between the receiver and the \textit{i}-th transmitter is given by:

\begin{equation}
    d_{i, \parallel} = \Big[{\frac{P_{t, i} A_{r} (m+1) h^{m+1}}{2 \pi P_{r,i}}}\Big]^{\frac{1}{(m+3)}}
\label{eq:distance_pr}
\end{equation}

With this method, a good accuracy can be obtained when A1 is true \cite{plets2019efficient}. However, the study of \cite{almadani2019novel} shows that when RSS is used to estimate height, the result is affected by tilt of the receiver. In turn, this has a considerable effect in the overall accuracy of the method.
%when tilt is introduced to the receiver, it has a detrimental effect on the estimation of height using RSS \cite{almadani2019novel}. 
%This considerably affects the overall performance of the method 
Thus, the \indirecth approach is inadequate when assumption A1 is broken. %If the wording of the previous two sentences needs to be changed it's ok, but please do not remove any words or information. I am trying to be very precise here and I believe every word matters.
%It is important that a 3D VLP method for drones performs well even when the receiver is tilted, since this occurs frequently during any flight, as explained in Sec.~\ref{sec:background}. 
We address the limitations of the \indirecth method to achieve a more accurate location estimation, even when A1 does not hold true. %In addition, we augment the common framework of RSS methods in our model to consider the effect of tilt of the receiver.

\subsection{Firefly}

%%%%%%%%%%% Previous draft
%In our approach, the violation of A1 is resolved by adopting a multi-sensor fusion approach. We design a robust height estimation method against tilting using both inertial sensors (IMUs) and barometric sensors. As most off-the-shelf drones are already equipped with IMUs and barometric sensors, the added hardware complexity is negligible. The violation of A2 is resolved by applying an additional iteration of trilateration, using direct height estimation and IMU data to account for tilt of the receiver, and thus provide a better estimation of the incidence and irradiance angles. Our proposed method consists of three steps:\\
%%%%%%%%%%%%%%%%%%%%%%%%%%%%%

In \name, we adopt a multi-sensor fusion approach. We design a \textit{direct} height estimation method that is robust against tilting using both inertial sensors (IMUs) and barometric sensors. % to overcome the limitations of the \indirecth approach. 
As most off-the-shelf drones are already equipped with %IMUs and barometric sensors,
both kinds of sensors, no additional hardware is needed.
%the added hardware complexity is negligible. 
We directly introduce our estimated height into the distance equations of the common framework and perform two iterations to tackle tilts in the irradiance and incidence angles. %to obtain an initial position via trilateration. Then we propose an additional iteration of trilateration, using IMU data combined with our previous results, to provide a better approximation of the incidence and irradiance angles. 
Our method consists of three steps:

\underline{\textit{Step 1:}} The height of the drone is measured using the onboard barometer and IMU with a complementary filter. The \indirecth method proposed in SoA studies is applied to correct the long-term zero drift of the barometer.

\underline{\textit{Step 2:}} We perform a \textit{first iteration} of the 2D trilateration method assuming that the transmitters and receiver are parallel. These (initial) location estimates tackle the tilt in the irradiance angle and serve as a basis for Step 3.

\underline{\textit{Step 3:}} Taking into account the tilting of the drone obtained from the IMU (incidence angle), we perform a \textit{second iteration} of the 2D trilateration method, which gives us the final localization results.

\begin{figure}[!t]
    \centering
    \includegraphics[width=\columnwidth]{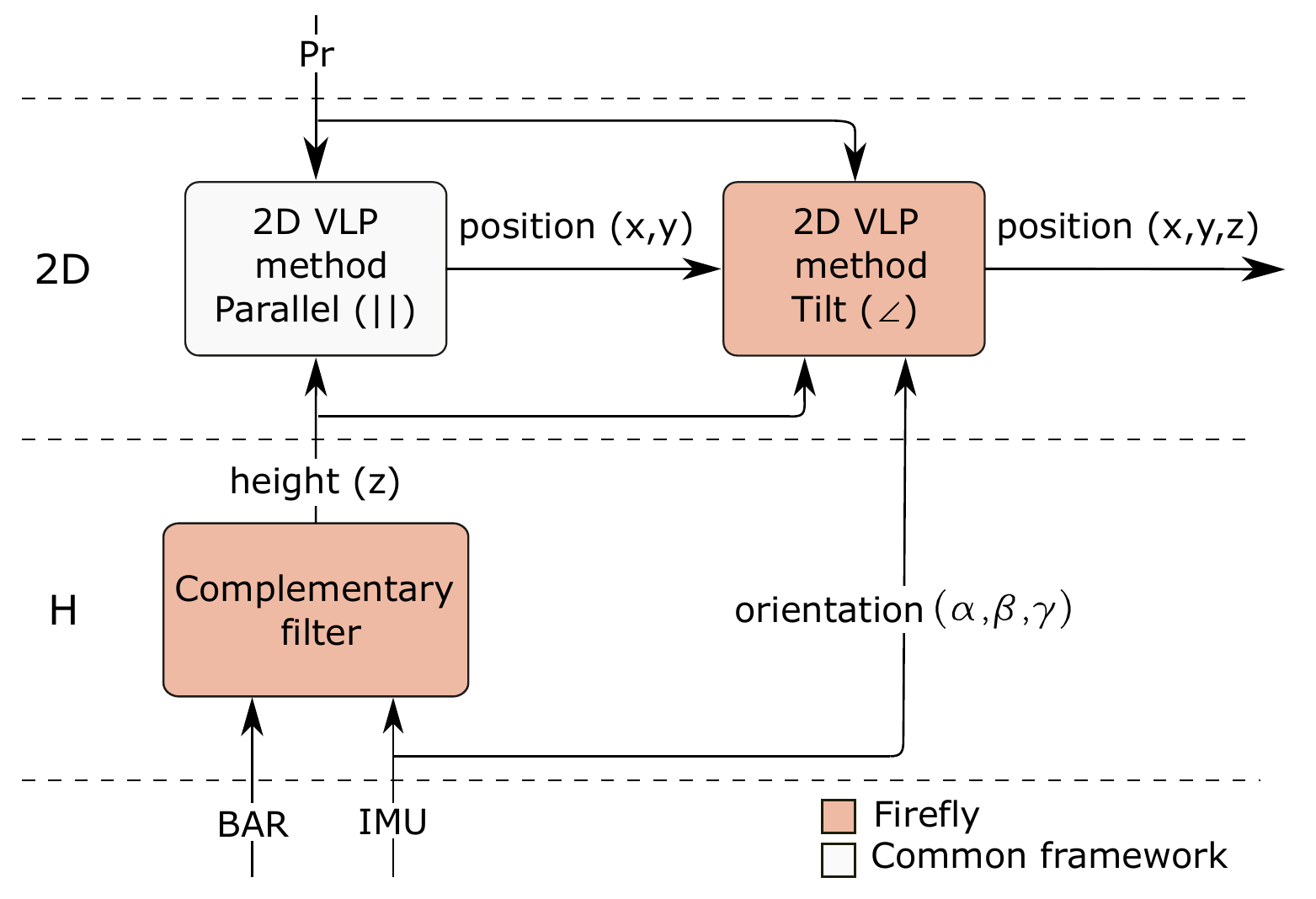}
    \caption{\name approach. The common framework of the SOA, which assumes a static parallel setup, is enhanced with two novel components: a direct measurement of height via a complementary filter and a 2D VLP method that considers tilt.}
    \label{fig:method_steps}
\end{figure}

\subsubsection{Step 1: Direct height estimation with sensor fusion}
\label{ss_sec:height_estimation}

To obtain a reliable measurement of the height of a drone, two types of sensors are often considered: IMUs and barometric sensors. IMUs consist of accelerometers and gyroscopes that are used to track the position and orientation of an object. Their low costs, form factors and power consumption make them an ideal option for drone navigation. While IMUs are good at measuring instantaneous changes, their outputs are prone to noise and drift errors, even over a short period of time~\cite{woodman2007introduction}. %By using only inertial sensors for positioning, it has been shown that the positioning error can reach over \SI{10}{\m} after \SI{2}{\s} and over \SI{150}{\m} after \SI{60}{\s}~\cite{woodman2007introduction, sabatini2014sensor}. 
On the other hand, the output of barometers, which use air pressure to measure altitude, is sensitive to rapid variations in the air pressure uncorrelated with altitude changes. %, which result in a very noisy output with many high-frequency noises in the short term. 
However, over a longer time period, barometers can provide a stable estimate of height.

Considering that the strengths of both kinds of sensors are complementary, %IMUs and barometric sensors are complementary, %where IMUs are more accurate in tracking instantaneous movements, and barometric sensors are more accurate in tracking long-term changes. Therefore, 
their respective advantages can be \textit{fused} to obtain an improved measurement using the \textit{long term} reference of the barometer and the \textit{short term} agility of the acceleration measurements. In Firefly, we adopt the complementary filter from \cite{higgins1974-complementary_vs_KF}, to %combine the measurements of the onboard IMU and barometer. %~\cite{higgins1974-complementary_vs_KF, sabatini2014sensor}.
%The complementary filter 
apply a high-pass filter to the vertical acceleration from the IMU and a low-pass filter to the output of the barometer. The complementary filter is especially suitable for a resource-limited UAV, compared to other sensor fusion methods such as the Kalman Filter, because of its simplicity.

\begin{figure}[t]
    \centering
    \includegraphics[width=2.5in]{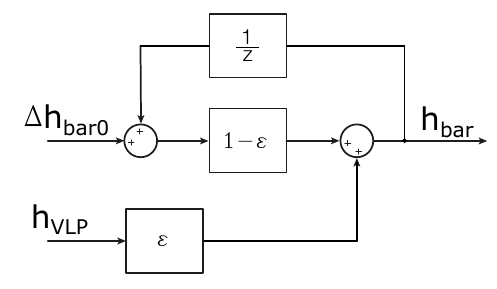}
    \caption{Block diagram for barometer drift correction scheme.}
    \label{fig:drif_correction}
\end{figure}

When combined with inertial sensors, barometers can provide accurate height estimation for drones, for example, with errors less than \SI{15}{\cm} after \SI{3}{\minute}~\cite{sabatini2014sensor}. %The barometer can provide a reliable long-term reference for the complementary filter within the battery time frame of drones. 
Nonetheless, due to changing conditions in the environment, barometers are likely to eventually drift from the initial conditions set during calibration. This is also known as \textit{zero drift}. One possible solution to correct the zero drift is to perform frequent calibrations at defined locations or checkpoints, but this approach is cumbersome. %In Firefly, we propose a more convenient solution for long term drift correction based purely on VLP.
In Firefly, we propose a simple zero drift correction scheme using VLP, which is shown in Fig.~\ref{fig:drif_correction}.

%However, due to the changing conditions of the environment, barometers are likely to eventually drift from the initial conditions set during calibration. This is also known as \textit{zero drift}. One possible solution to correct the zero drift is to perform frequent calibrations at defined locations or checkpoints, but this approach is cumbersome. In Firefly, we propose a more convenient solution for long term drift correction based purely on VLP.

As described in Sec.~\ref{sec:SoA-2dIh}, the \indirecth method already provides a solution to indirectly estimate the height of a drone using VLP. This solutions is not accurate when tilting occurs, but provides a drift-free height measurement. Thus, we use the \indirecth measurement to correct the drift problem, but only when it can be trusted, i.e. when the roll and pitch angles of the drone are below a threshold such as 3$^\circ$ using information from the IMU. 

%We have mentioned that this method works well in static conditions but is not accurate when tilting occurs. Although we intend to overcome the limitations of this approach in a mobile scenario by measuring the height directly with IMUs and barometers, the \indirecth methods provides a drift-free height measurement that is leveraged to overcome the zero drift problem. 

In our method, the height estimated by the \indirecth VLP is denoted as $h_{VLP}$, and the change of height measured by the barometer is denoted as $\Delta h_{bar0}$. Overall, $\Delta h_{bar0}$ has a higher accuracy than $h_{VLP}$ to detect changes in altitude and we show this in \cref{ssec:experimental_setup}. 
The measurement results $h_{VLP}$ and $\Delta h_{bar0}$ are then scaled by a constant $\epsilon$ and $1-\epsilon$, respectively, to obtain the final height estimation $h_{bar}$. In the following iterations, $h_{bar}$ then acts as a correction for $\Delta h_{bar0}$. By choosing a small value for the constant $\epsilon$, the \indirecth height estimation ($h_{VLP}$) has a minor influence in the immediate output. This ensures that the height estimation is still accurate. At the same time, the drift-free $h_{VLP}$ has an anchor effect on the long term, which corrects zero drift from the barometer. Together with the IMU data, $h_{bar}$ is the input to the complementary filter.

\subsubsection{Step 2: Parallel TX-RX}
%\subsubsection{Step 3: Irradiance angle}
\label{ss_sec:parallel-tx-rx}

In the previous step, we directly obtain the height of the receiver ($z_r$). Now, we utilize this accurate height to obtain a first estimation of the $x_r$ and $y_r$ coordinates of the receiver. This step tackles the issue caused by the incorrect characterization of the irradiance angle in \indirecth. By directly estimating height, we obtain a better approximation of the irradiance angle than the \indirecth method.
For this first iteration, we consider the common framework of RSS methods where the receiver is parallel to the transmitter. We take into account the effect of the tilt of the drone (incidence angle) in the next step. %(\cref{ss_sec:tilted-tx-rx}).

In a  typical  indoor  environment  the transmitters are fixed, and the irradiance angle $\psi$ is not affected  by  tilting  of  the  receiver.  Then,  for  the  irradiance angle, \cref{eqn:parallel_assumption} is valid and we can use \cref{eq:distance_pr} to obtain distance, provided that the height estimation is accurate. Note in \cref{fig:drone_ref} that given an estimation of $d_i$ (based on RSS measurements), a correct height is critical to obtain a proper characterization of $\psi$, and that is the strength of \name compared to \indirecth. %In the \indirecth approach we have mentioned that the height estimation is not accurate when the receiver is tilted. In our method, we have directly estimated the altitude of the drone in \cref{ss_sec:height_estimation} to overcome this limitation.

With the distance estimation we can obtain the position of the drone using a standard trilateration procedure. In our system, as we are using visible light for positioning, we can consider the main sources of noise to be the shot and thermal noises \cite{hua2018-noise-vlp}. Therefore, we adopt the maximum likelihood estimation (MLE) to perform 2D trilateration, as it takes into account statistical considerations to minimize the error of noisy measurements~\cite{goldoni2010-rssi-indoor-loc}.\\

\subsubsection{Step 3: VLP with tilting}
%\subsubsection{Step 3: Incidence angle}
\label{ss_sec:tilted-tx-rx}
In the previous step, we obtain the position of the receiver assuming that the incidence and irradiance angles are equal. As we explained before, the irradiance angle is not affected by tilting in our setup. On the other hand, the incidence angle is directly influenced by the 3D orientation of the receiver. We can obtain this information using the IMU which we already use to estimate the height of the drone. In this step, we perform an additional iteration of the 2D MLE method, using distance measurements that account for tilt ($\angle$). From equation \cref{eq:rxowerp_w_nf} we have:

%Given the direct height estimation of step 1 (\cref{ss_sec:height_estimation}), the distance obtained assuming parallel irradiance and incidence angles is accurate a close approximation since the height estimation was not deteriorated by tilting of the receiver.

%accurate height estimation that is not affected by tilting and thus we obtain a close estimation of distance using. However,  we are still assuming that the receiver and transmitter are parallel to each other. As we explained before, this assumption will be frequently incorrect when the drone is flying. In a typical indoor environment the transmitters are fixed, and the irradiance angles are not affected by tilting. On the other hand, the incidence angle is directly influenced by 3D orientation of the receiver. We can obtain this information using the IMU which we have already used to estimate height of the drone. For this reason, we perform an additional iteration of the 2D MLE method to account for tilt ($\angle$). From equation \cref{eq:rxowerp_w_nf} we have that:

\begin{equation}
    d_{i, \angle} = \sqrt{\frac{P_{t,i} A_{r} (m+1) cos^m(\psi)cos(\theta)}{2 \pi P_r}}
\end{equation}
where the irradiance angle $\psi$ is described by equation \cref{eqn:parallel_assumption} and the incidence angle $\theta$ is described by \cite{yang2014three}:

%\begin{strip}
% \begin{equation}
% \theta = cos^{-1} \frac{(x_r-x_i)cos(\gamma)sin(\beta)+(y_r-y_i)sin(\gamma)sin(\beta) + (z_r-z_i)cos(\beta)}{d_{i, ||}}
%\label{eq:tilt_receiver}
%\end{equation}
%\end{strip}

 \begin{equation}
 \theta = cos^{-1} \Big( \frac{\theta_x + \theta_y + \theta_z}{d_{i, ||}} \Big)
\label{eq:tilt_receiver}
\end{equation}
with
\begin{center}
$ \theta_x = (x_r-x_i)cos(\gamma)sin(\beta)$\\
\vspace{2mm}
$\theta_y = (y_r-y_i)sin(\gamma)sin(\beta)$\\
\vspace{2mm}
\hspace*{-1cm}$\theta_z = (z_r-z_i)cos(\beta)$\\
\end{center}
where $\gamma$, $\beta$ are the roll and pitch Euler angles respectively obtained from the IMU. We can see from \cref{eq:tilt_receiver} that the first iteration of the 2D trilateration method (Step 2) is necessary since the position of the receiver ($x_r$, $y_r$ and $z_r$) is required in addition to its orientation for $\theta_x$ and $\theta_y$.

\section{Experimental Validation} %Ricardo
\label{sec:Experimental_Validation}
\begin{comment}
In this section, we describe the system evaluation carried out in an indoor testbed with a flying drone with 6 DOF. %The overview of the setup is shown in Fig.~\ref{fig:eval}. 
\end{comment}

In this section, we evaluate Firefly and compare it with SoA methods in an indoor testbed shown in Fig.~\ref{fig:eval}. We select two methods as reference: the \textit{3D VLP with RSS} method in \cite{carreno20-metaheuristic} and the \textit{2D+H VLP with RSS} method in \cite{plets2019efficient} which have reported favorable results, with errors below 10\,cm, as previously shown in \cref{fig:soa_graph}.
%We select two methods as references: the \textit{3D VLP with RSS} method in \cite{carreno20-metaheuristic} and the \textit{2D+H VLP with RSS} method in \cite{plets2019efficient} which have reported favorable results with static and parallel receivers. %Note: the methods implemented were tested in simulation. But very similar alternatives were tested experimentally with comparable results in \cite{cai2017-vlp-pso} and \cite{almadani2019novel} respectively, also using static and parallel receivers. 
Two metrics are used for comparison: 1) the location error, calculated as the Euclidean distance from ground truth to the estimation of each method, and 2) the algorithm complexity.  

\subsection{Overview of the testbed}

\begin{figure}[!t]
\centering
\subfloat[]{
    \includegraphics[width=0.49\columnwidth]{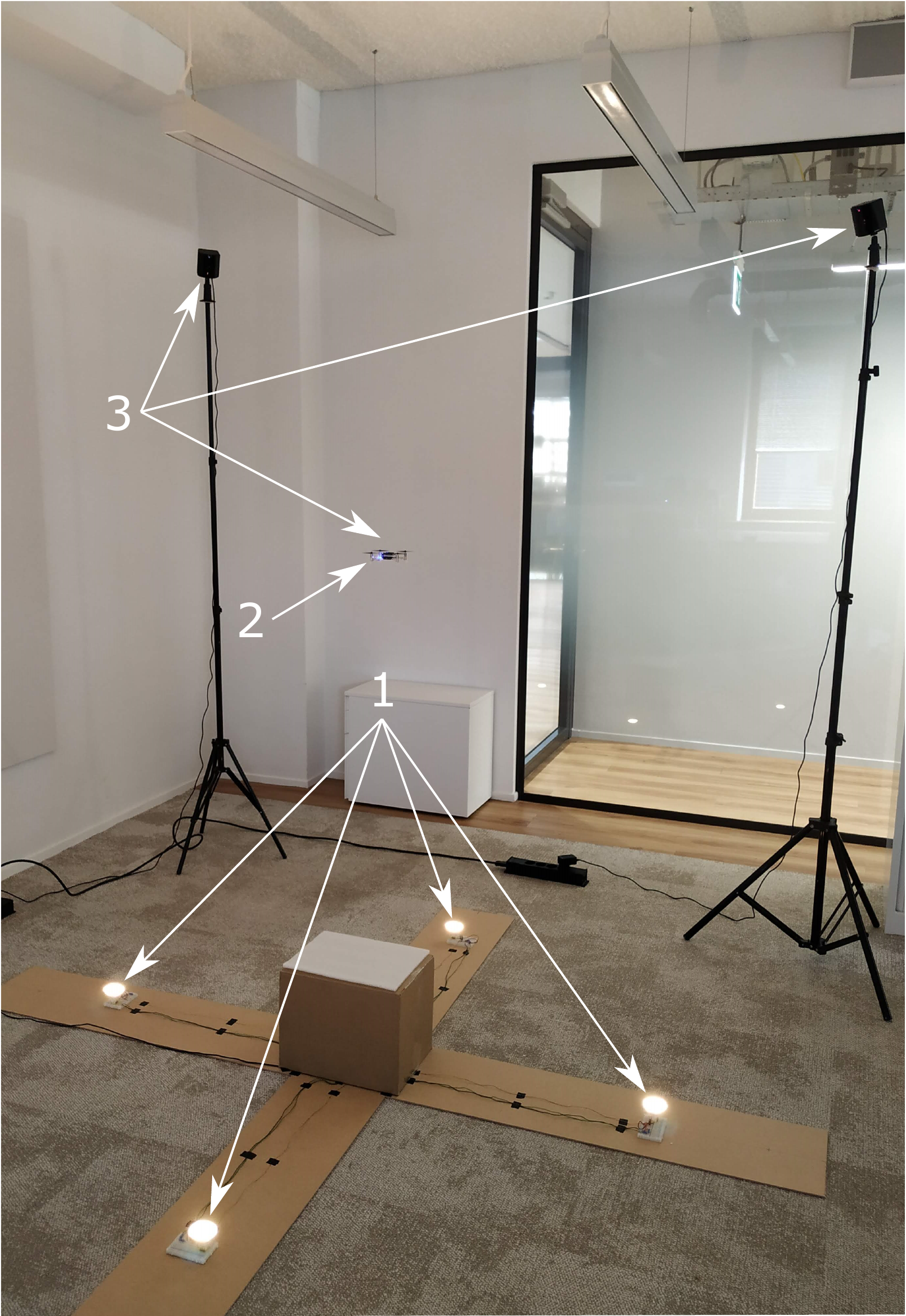}
    \label{fig:test_setup}}
\subfloat[]{
    \includegraphics[width=0.51\columnwidth]{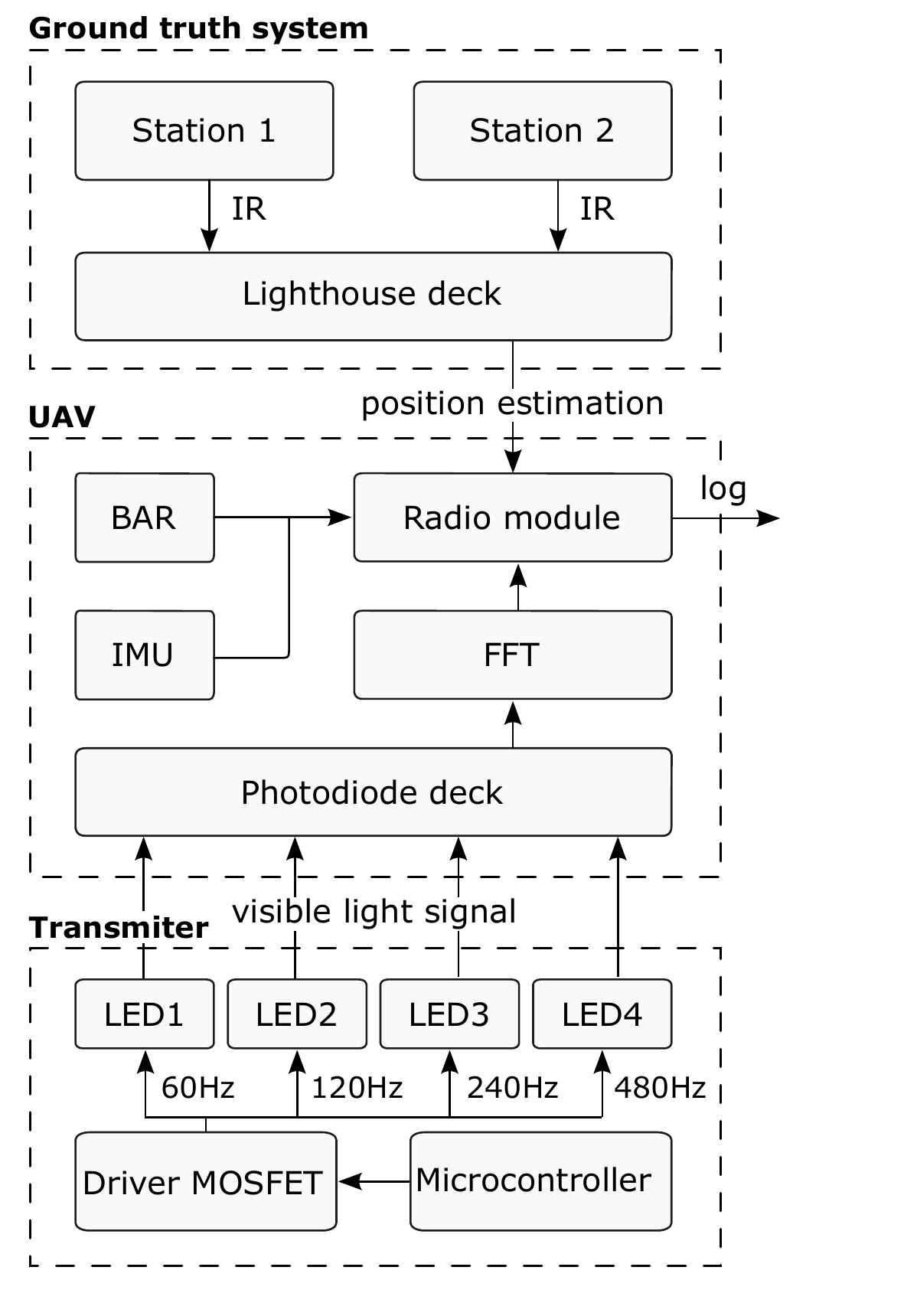}
     \label{fig:Implementation}}
    \caption{The overview of the system. (a) The physical position of the testbed with 1) 4 transmitters; 2) UAV; and 3) the ground truth system.
    (b) Schematic overview of the connection and functionalities of the components.} 
    \label{fig:eval}
\end{figure}

We build a testbed in a 2 m x 2 m x 2 m indoor environment as shown in \cref{fig:test_setup}. The testbed has three main components: 1) 4 transmitting LED sources located at fixed positions, 2) the drone, and 3) a ground truth system to determine and control the drone's position during flight. The connection between these components and the schematic overview of their functionalities are shown in \cref{fig:Implementation}. The configuration of the system is listed in \cref{tab:sys_parameters}.

% Please add the following required packages to your document preamble:
% \usepackage{multirow}
\begin{table}[]
\centering
\begin{tabular}{|l|l|c|l|}
\hline
\multicolumn{1}{|c|}{\textbf{Parameter}} &
  \multicolumn{1}{c|}{\textbf{Label}} &
  \multicolumn{2}{c|}{\textbf{Value}} \\ \hline
Testbed size           & -            & \multicolumn{2}{c|}{2 m x 2 m x 2 m} \\ \hline
\multirow{4}{*}{\begin{tabular}[c]{@{}l@{}}Position (x, y, z)  and \\ frequency ID of each\\ transmitter\end{tabular}} &
  $Tx_1$ &
  (\SI{0.25}{\m}, \SI{1}{\m}, \SI{0}{\m})&
  \SI{60}{\hertz} \\ \cline{2-4} 
                       & $Tx_2$       & (\SI{1}{\m}, \SI{1.75}{\m}, \SI{0}{\m})   & \SI{120}{\hertz}  \\ \cline{2-4} 
                       & $Tx_3$       & \SI{1.75}{\m}, \SI{1}{\m}, \SI{0}{\m})    & \SI{240}{\hertz}  \\ \cline{2-4} 
                       & $Tx_4$       & (\SI{1}{\m}, \SI{0.25}{\m}, \SI{0}{\m})   & \SI{480}{\hertz}  \\ \hline
Transmitter power      & $P_t$        & \multicolumn{2}{c|}{\SI{4.7}{\W}}                             \\ \hline
Lambertian order       & $m$          & \multicolumn{2}{c|}{14}                                       \\ \hline
Area of the photodiode & $A_{r}$     & \multicolumn{2}{c|}{\SI{5.2}{\mm\squared}}                    \\ \hline
FOV of the receiver    & $\Theta_{c}$ & \multicolumn{2}{c|}{\SI{160}{\degree}}                        \\ \hline
\end{tabular}
\caption{Parameters of the system. }
\label{tab:sys_parameters}
\end{table}

\subsubsection{Transmitter}
CorePro LEDspot LV lamps are used as transmitters. Each LED lamp is controlled by a microcontroller to generate a unique frequency identifier (ID) which will be explained in \cref{ssec:modulation}.

\subsubsection{Receiver}
\label{sec:Receiver}
%Drone platform, PCB, sensor, ground truth. Explanation of setup.\\

\begin{figure}[!t]
    \centering
    \subfloat[]{
    \includegraphics[width=0.45\columnwidth]{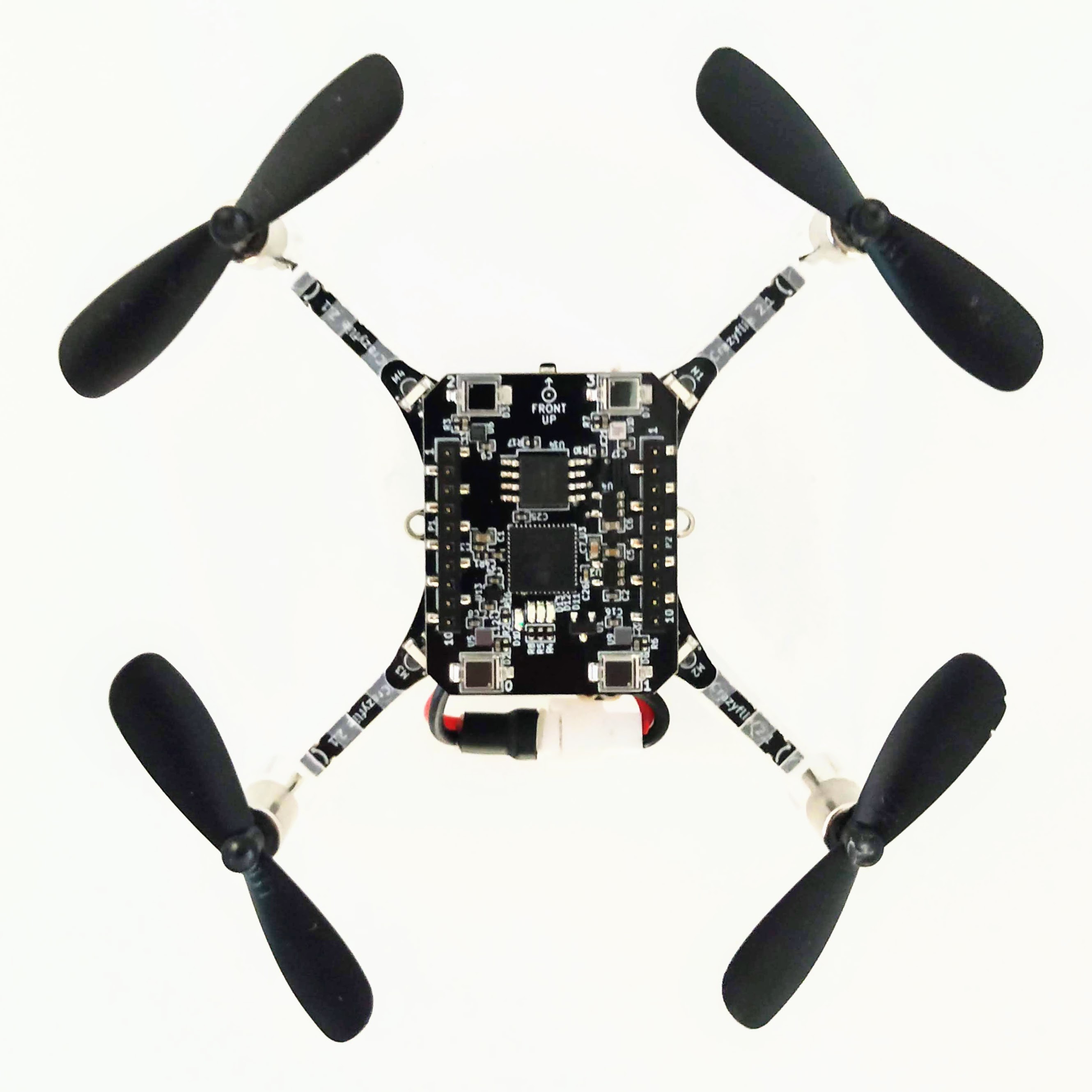}
     \label{fig:cf_decks_lh}
    }
    \subfloat[]{
    \includegraphics[width=0.45\columnwidth]{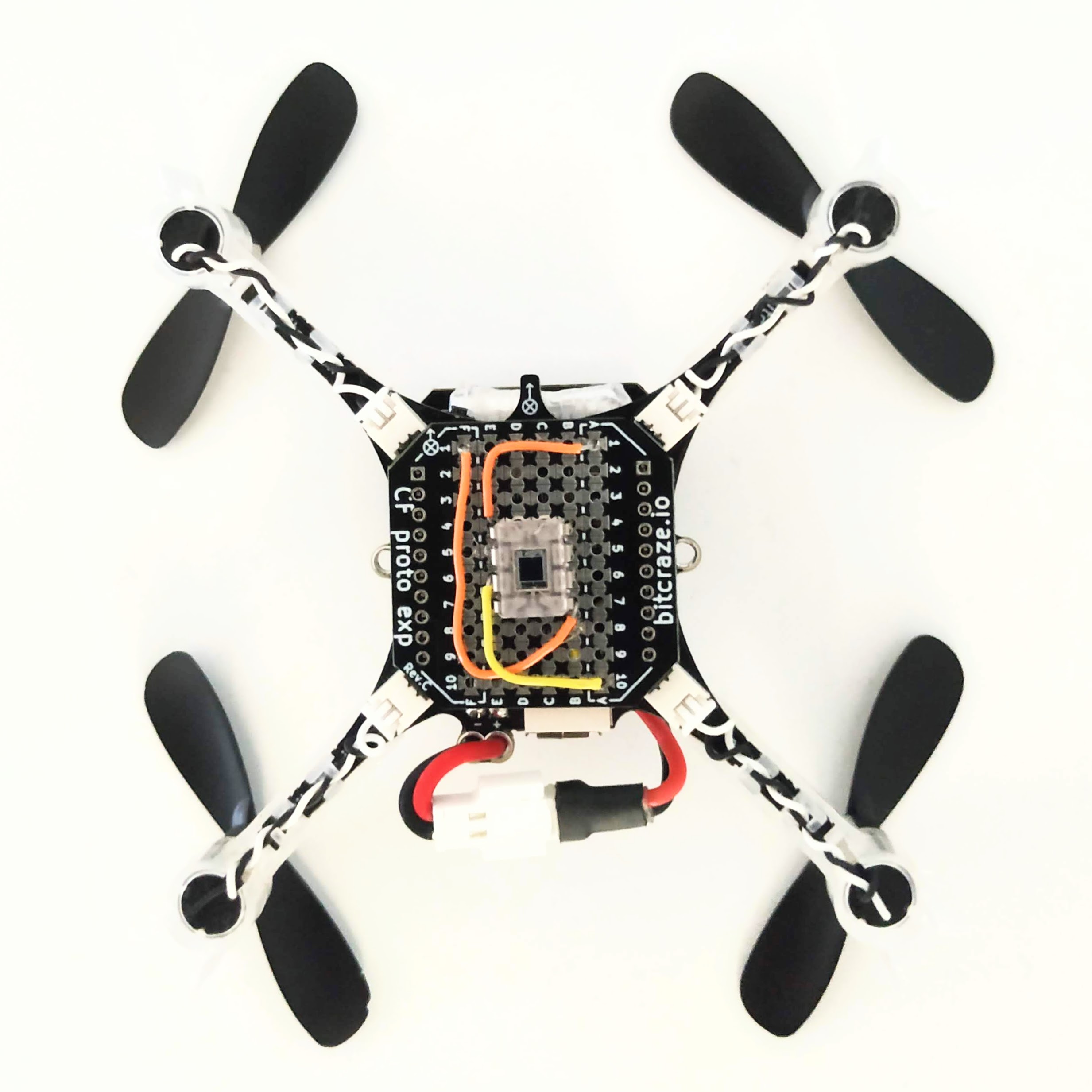}
     \label{fig:cf_decks_pd}
    }
    \caption{Crazyflie 2.1 mini-drone (a) Lighthouse deck used for ground truth (top view) and (b) custom PD deck (bottom view)}
\end{figure}

We select the Bitcraze Crazyflie 2.1 as the receiver. It is a lightweight commercially available drone with multiple peripherals, which makes it easy to attach dedicated or customized hardware to perform different functions. To enable VLP on the drone, an OPT101 PD is attached via a custom PCB as shown in \cref{fig:cf_decks_pd}. The PD is connected to a 12-bit ADC on the Crazyflie. %We transmit all of the sensor data (\todo{check this} from the drone to a remote client using the onboard \SI{2.4}{GHz} radio module.

%Benchmark of how much the library can improve performance ~20-30x

\subsubsection{Ground truth system}
We use the Lighthouse positioning system from Bitcraze to control and retrieve the ground truth position of the UAV. It consists of two parts: the Steam VR Stations positioned above the flying range of the receiver and the positioning deck attached to the top of the Crazyfile, as shown in \cref{fig:cf_decks_lh}. The ground truth system uses two infrared (IR) stations to locate the positions of the drone and provides a high accuracy of less than \SI{10}{\cm}. However, these dedicated hardware stations 1) are more expensive (more than \$200 USD per station compared to less than \$5 USD for an LED lamp); %2) consume higher power (a single station consumes 10W), 
2) need careful setup and calibration; and 3) cannot utilize the existing indoor illumination infrastructure compared to Firefly. Thus, the chosen ground truth system is costly and impractical for drone positioning in a large scale application. However, it is an effective research tool to benchmark the performance of Firefly and SoA methods.

%\subsection{Visible light channel}
\subsection{Multiplexing}
\label{ssec:modulation}
In Firefly, each LED transmits a unique frequency ID in the form of a square wave by turning itself on and off. In order to access the shared medium, the four LEDs use frequency division multiple access (FDMA) to transmit their unique IDs simultaneously. The combined signal is sampled by the PD in the receiver, and then decomposed by a Fast Fourier Transform (FFT) to obtain the RSS ($P_r$) corresponding to each frequency ID (i.e. the light source). 
In this way, Firefly is resilient to the presence of high frequency shot and thermal noise as well as constant DC component of ambient light. This allows \name to work in dark and illuminated environments with a simple implementation (i.e. no synchronization and no additional protocol is required by FDMA). To showcase its robustness, our experiments were carried under the presence (interference) of externals sources of artificial and natural light.
%With FDMA, the available frequency spectrum is divided into separate frequency bands to simultaneously transmit information. 
%It is simple to implement because no synchronization is needed.
%it allows the LEDs to transmit their IDs simultaneously without the need to be synchronized. FDMA is also efficient since no communication protocol needs to be enforced when transmitting only the frequency ID. 
%Using FDMA for our method offers significant advantages in reliability, as the measurements will be resilient to sources of high frequency noise present and the constant DC component of ambient light. This will allow \name to work in dark and illuminated environments.

%In our system, each LED transmits a unique frequency ID in the form of a square wave. The resulting signal is sampled by the PD in the receiver, and then decomposed using a Fast Fourier Transform (FFT), which allows us to obtain the RSS from distinct light sources. 

Note that, the frequency IDs of LEDs in Firefly are selected to avoid interference. As the Fourier transformation of a periodic square signal contains odd-harmonic components, using $2^n$ multiples of a selected base frequency (i.e. $2f_0, 4f_0, 8f_0, etc.$) can resolve the interference problem ~\cite{delausnay2015-fdmsquarewaves}. The base frequency in Firefly is arbitrarily chosen to be \SI{60}{\hertz} for evaluation. However, any base frequency can be used, as long as the hardware limitations of the receiver and transmitters are taken into account.
%When select the frequency IDs of LEDs, it is important to consider that for a periodic square signal, the output of its FFT will contain odd-integer harmonic components. Therefore, we choose transmitting frequencies that are $2^n$ multiples of a selected base frequency (i.e. $2f_0, 4f_0, 8f_0, etc.$) to avoid interference~\cite{delausnay2015-fdmsquarewaves}. In our system, we arbitrarily choose the base frequency to be \SI{60}{\hertz} for evaluation. However, any base frequency can be used, as long as the hardware limitations of the receiver and transmitters are taken into account.

\subsection{Evaluation results}
\label{ssec:experimental_setup}

\begin{comment}
We evaluate Firefly and compare it with SoA methods in an experimental setup with a mobile receiver subject to 6 DOF. To the best of our knowledge, this is the first time VLP has been empirically tested under such demanding conditions. We select two methods as references: the \textit{3D VLP with RSS} method in \cite{carreno20-metaheuristic} (similar to \cite{cai2017-vlp-pso}) and the \textit{2D+H VLP with RSS} method in \cite{plets2019efficient} which have reported favorable results with static and parallel receivers. %Note: the methods implemented were tested in simulation. But very similar alternatives were tested experimentally with comparable results in \cite{cai2017-vlp-pso} and \cite{almadani2019novel} respectively, also using static and parallel receivers. 
The Euclidean norm is used to calculate the distance error between the ground truth and the estimation of each method.
\end{comment}
We carry out 8 automated flight tests to cover a circular path of radius \SI{0.5}{\m} and a range of heights from \SI{0.2}{\m} to \SI{1.8}{\m}. In each flight test, the drone follows a pre-programmed route with curves and direction changes. As a result, we are constantly inducing tilting on the drone to test the VLP methods in a real flight scenario. Along the trajectory, the drone is sampling the visible light signal and executes the FFT in real time to retrieve the ID and the received power from each LED source.
\begin{comment}
While our approach can be run real-time in the drone during tests, to execute the three algorithms and compare results for the different approaches would be too demanding to achieve in real-time. Therefore, we send the sensor data log to a remote server and execute the selected algorithms as shown in \cref{fig:Implementation}. The data transmitted from multiple sensors of the drone to the remote server is shown \cref{tab:log_var}. 
\end{comment}
Although Firefly can run in real time in the drone, executing the three algorithms simultaneously is too demanding. Therefore, we send the sensor data log to a remote server as shown in \cref{fig:Implementation} and use Matlab to compare them. \cref{tab:log_var} lists the data transmitted from the drone. 

\begin{table}[]
\centering
\begin{tabular}{|l|l|l|l|}
\hline
\textbf{Source}      & \textbf{Variable (s)} & \textbf{Units} & \textbf{Description}     \\ \hline
PD                   & $Pr 1-4$              & [$\mu W$]      & Power received           \\ \hline
\multirow{2}{*}{IMU} & $a_x, a_y, a_z$       & [Gs]           & Linear acceleration      \\ \cline{2-4} 
                                                           & $\alpha, \beta, \gamma$ & [ $^\circ$ ] & \begin{tabular}[c]{@{}l@{}}Orientation \\ (roll, pitch, yaw)\end{tabular} \\ \hline
Barometer            & $h_{bar0}$            & [m]            & Altitude above sea level \\ \hline
\begin{tabular}[c]{@{}l@{}}Lighthouse \\ deck\end{tabular} & $x,y,z$                 & [m] & Ground truth position                                                     \\ \hline
\end{tabular}
\caption{Sensor variables transmitted to the remote client.}
\label{tab:log_var}
\end{table}

% Although the steps of the method after the FFT such as the distance estimation, MLE, sensor fusion and tracking could be performed online inside the drone while meeting the real-time requirements of the application, in order to facilitate the collection and presentation of statistical results when comparing the results of different methods (which overload the memory capabilities of the UAV), it was chosen to partly process the remaining steps outside the receiver. 

% A schematic showing the implementation of the system for the tests performed is shown in \cref{fig:Implementation} and the parameters of the system that were used for the calculations in the experimental tests are summarized in \cref{tab:sys_parameters}.\\
% \begin{figure}[!t]
%     \centering
%     \includegraphics[width=3.5in]{figures/implementation2.png}
%     \caption{Schematic of the implemented system.  \todo{(To be re-drawn in vector format) when you redraw this figure, maybe put the name of each component within the dash block? add the parameters besides the connection arrow of two components, e.g. from transmitter to UAV, add `60hz light source' etc., from Ground truth system to UAV, add IR?  }}
%     \label{fig:Implementation}
% \end{figure}

% \subsection{Results}
% \label{section:Results}

%\subsection{Evaluation Results}
%\label{ss:Results}

\subsubsection{Positioning error}
\begin{figure}[t]
    \centering
    \includegraphics[width=\columnwidth]{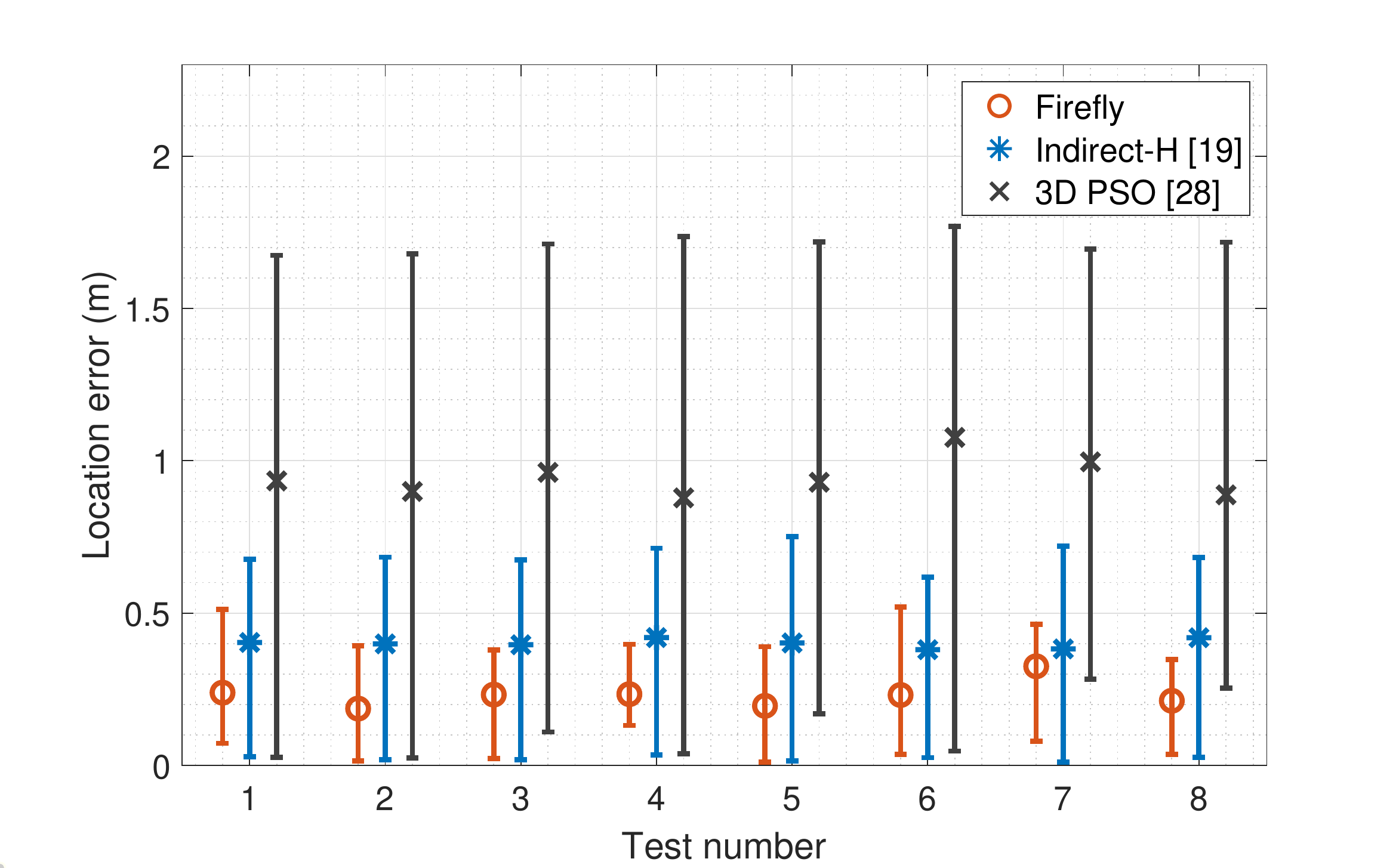}
    \caption{Mean-error plot for the 8 test flights.}
    \label{fig:Mean_error_plot}
\end{figure}

\cref{fig:Mean_error_plot} shows the mean-error of three methods across the 8 different tests. We label the method from \cite{plets2019efficient} as \textit{\indirecth} and the method from \cite{carreno20-metaheuristic} as \textit{\pso}. \name clearly outperforms the other two algorithms, in terms of the mean and maximum error achieved for all tests, by almost a factor of $\times 2$ and $\times 5$ with respect to \indirecth and \pso, respectively. \name and \indirecth are closer in terms of accuracy compared to \pso which evidently displays the lowest performance of all. Notice that in \cref{fig:soa_graph}, the results reported by \indirecth and \pso are below \SI{0.1}{m}, but that performance reduces significantly when tested in a mobile scenario with 6 DoF, obtaining errors around \SI{0.4}{m} and \SI{0.9}{m}, respectively. We will now discuss in detail the results obtained in \name with respect to each method.

\begin{figure}[t]
    \centering
    \includegraphics[width=0.9\columnwidth]{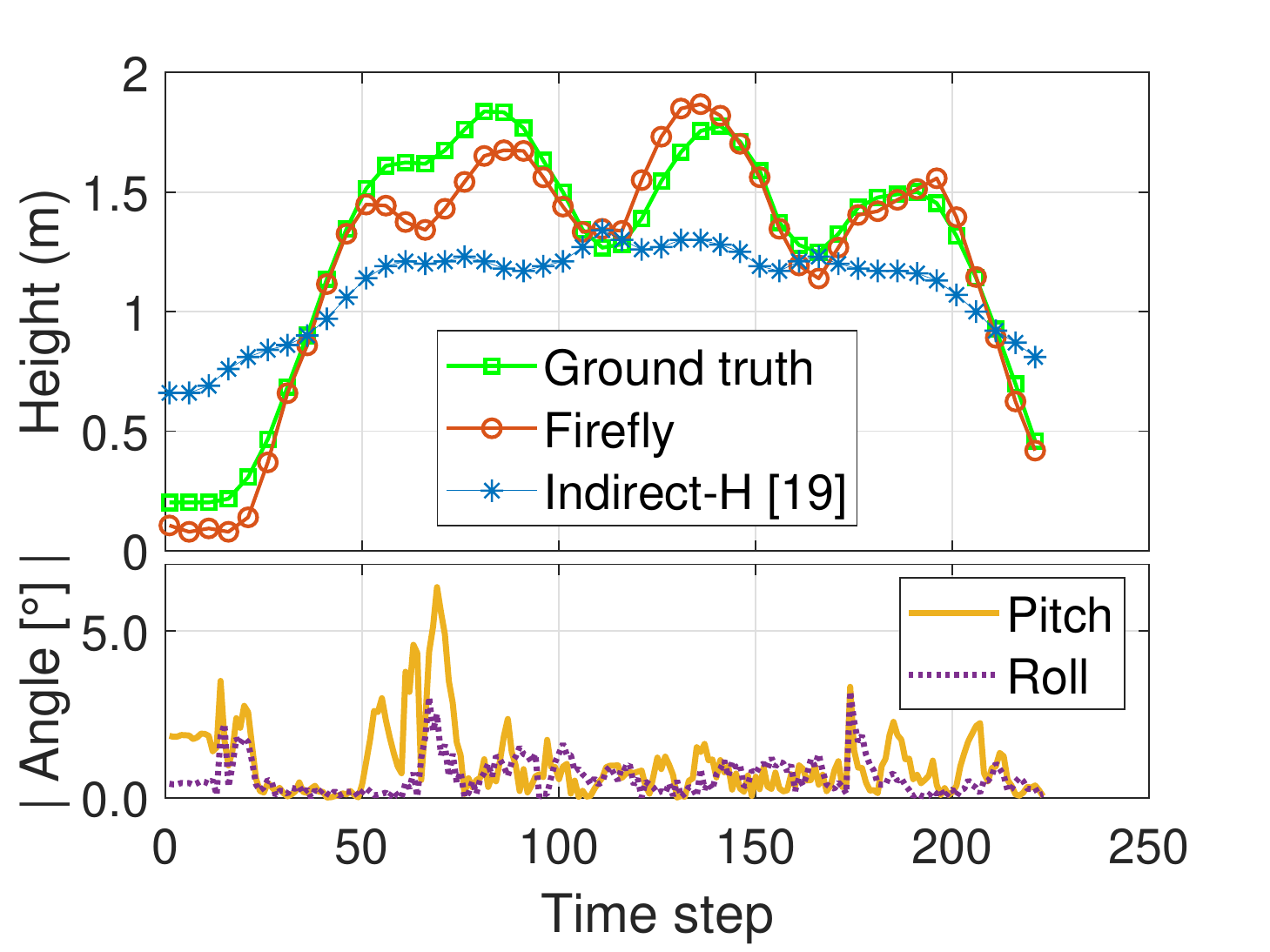}
    \caption{Height estimation (top) and tilting angles of the receiver (bottom) in Test 5.}
    \label{fig:route_comparison_2d_h}
\end{figure}

\underline{\textit{\name Vs. \indirecth:}} Let us take a close look of the \indirecth results in one of the tests. In \cref{fig:route_comparison_2d_h}, we show the height estimation of \indirecth and \name, and the receiver angles during the flight test. During the lift-off (timestep 0 to 60) and landing (after timestep 200), \indirecth is able to detect the monotonic changes in altitude, but to a much smaller extend compared to \name. More significantly, during the route (timestep 60 to 200), \indirecth fails to capture the variation in the height. In contrast, \name closely resembles the profile of the ground truth, and does not show any significant drift effect over the whole flight test. 

Considering the height measurements of all tests, the mean absolute error for \name is less than \SI{14}{\cm} compared to a mean error of \SI{30}{\cm} of the \indirecth method. In our tests, the drone is exposed to tilting angles up to 7$^\circ$  as shown in \cref{fig:route_comparison_2d_h} (bottom). Although \textit{some} of the errors of the \indirecth method can be attributed to tilting, the height estimation does not accurately reflect the ground truth even when the tilting angle of the receiver is small. For example, in the segment between timestep 100 and 150, the roll and pitch angles are less than \SI{2}{\degree}. Even though the receiver and transmitter are close to parallel during this time, we can see that the estimated height by the \indirecth method does not match to the ground truth from \cref{fig:route_comparison_2d_h} (top). Our approach considers the input of the indirect method but only with a small weight $\epsilon$ to compensate drifts, c.f.~\cref{fig:drif_correction}. %We speculate that the accuracy of the \indirecth method is very susceptible to minor tilting and tilting oscillations that are present in our mobile tests, but further investigation is required. 
As a result, \name produces an accurate height estimation using multi-sensor fusion.

\begin{figure*}[h]
    \centering
    \subfloat[3D Trajectory]{
    \includegraphics[width=0.46\textwidth]{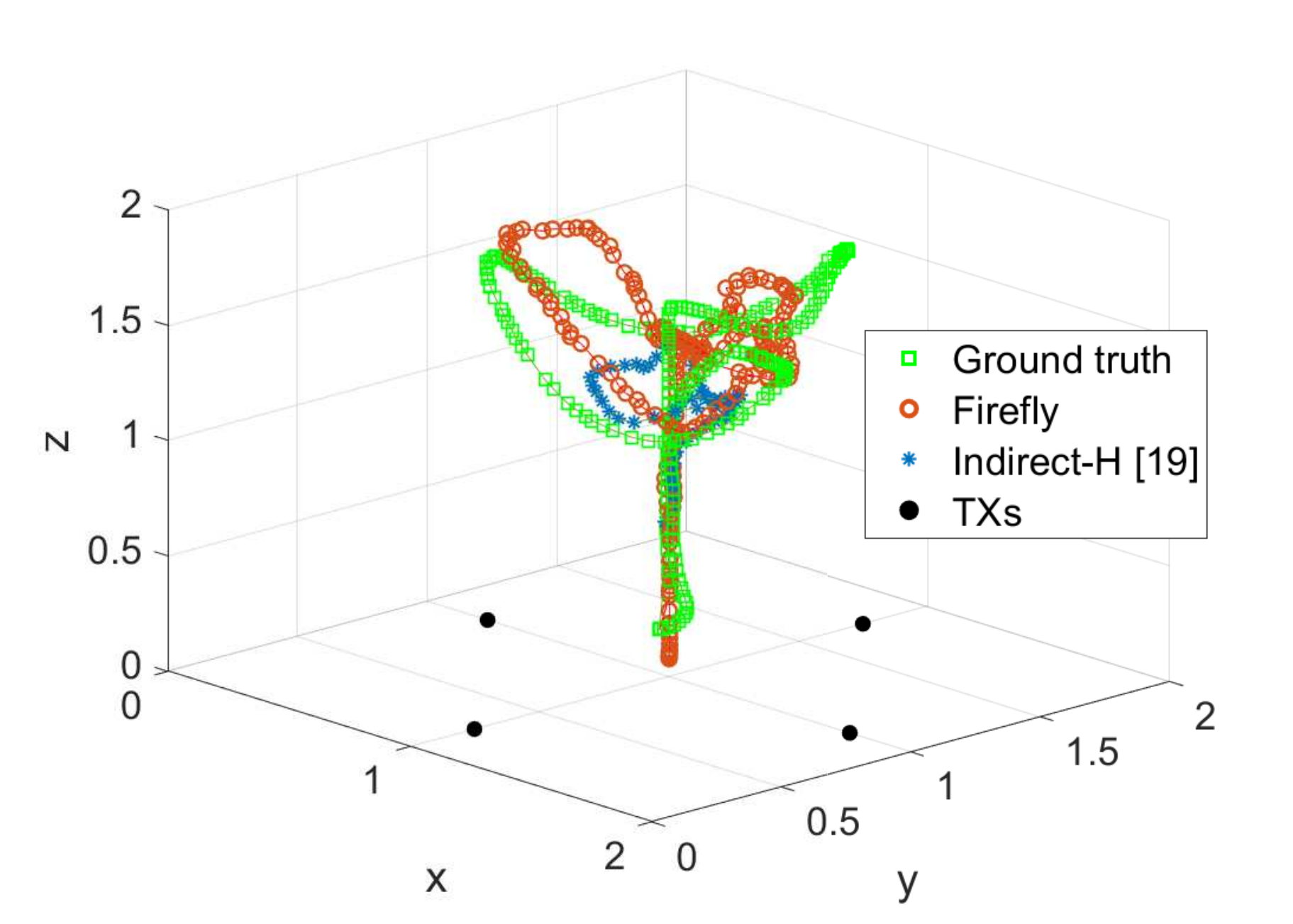}
    \label{fig:route_comparison_2d}
    }
   \subfloat[3D Trajectory (Top View)]{
    \includegraphics[width=0.46\textwidth]{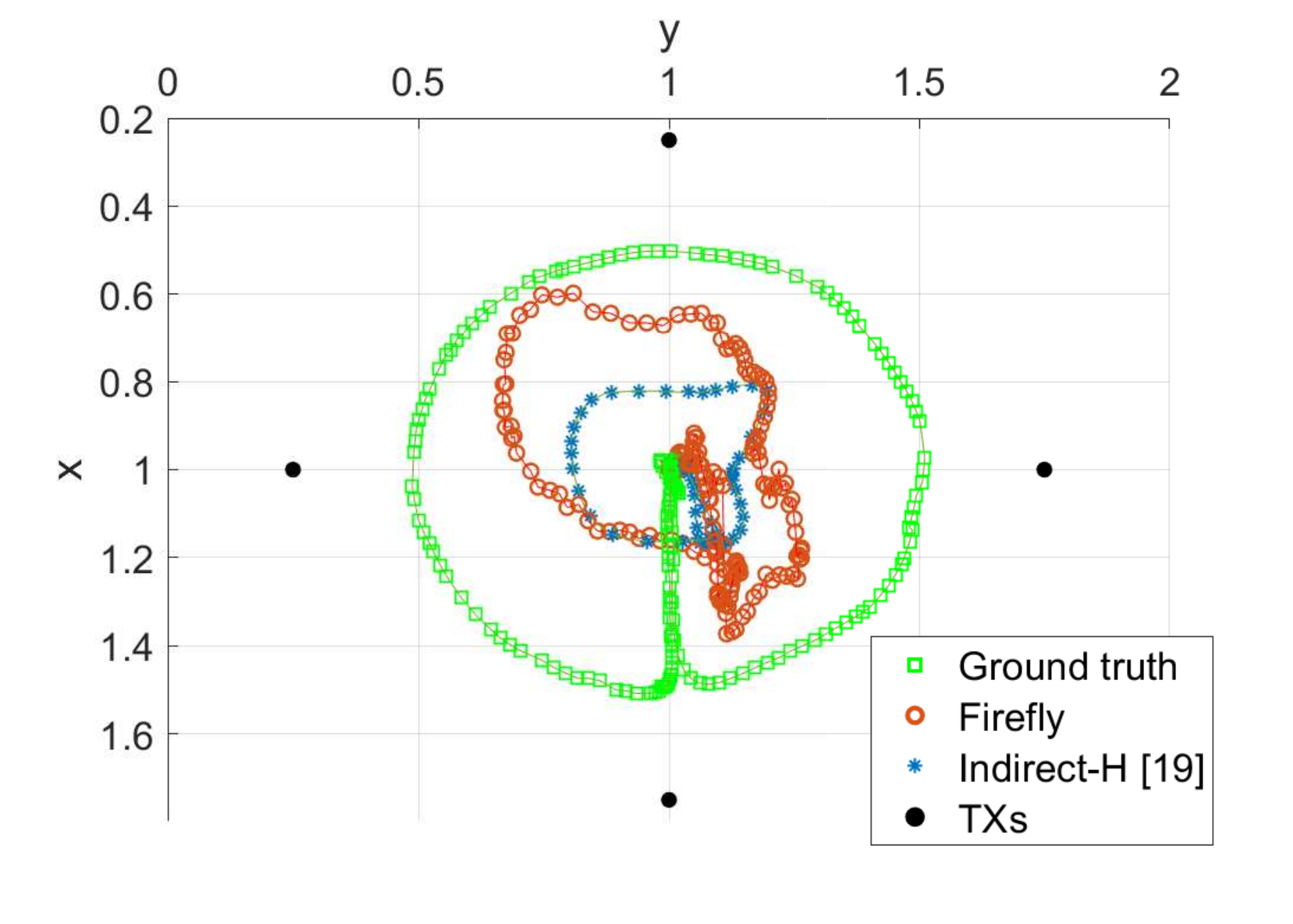}
    \label{fig:route_comparison_2d_top}
    }
     \caption{\name vs. \indirecth \cite{plets2019efficient} in Test 5.}
    \label{fig:firefly_vs_indirect}
    \vspace{-0.7cm}
\end{figure*}

In \cref{fig:route_comparison_2d_top}, we now analyze the 2D position by looking at the top view of the trajectory for the same test. The \indirecth estimation roughly captures the circular motion of the actual drone, but not with the amplitude seen in the ground truth. \name achieves a wider amplitude in the x-y plane that is closer to ground truth. Since the 2D position of the receiver is computed with the same RSS information in both methods, this improvement of the 2D position can be mainly attributed to (1) an accurate height estimation using a sensor-fusion approach and (2) the consideration of tilting of the receiver into the equations. 

\cref{fig:route_comparison_2d} depicts the 3D trajectory of both methods. It supports our previous analysis that \name provides a more accurate location than \indirecth. From the statistical measurements of the position error of all 8 tests, \name improves the accuracy by around 42\% as listed in \cref{tab:Pos_accuracy_improvement}.
%we present statistical measurements of the position error and the improvement of \name compared to the \indirecth method, which is the closest in terms of performance. To obtain this results, we used the error measurements considering all 8 tests. \\

\vspace{0.2cm}

\begin{table}[h]
\centering
\begin{tabular}{|l|l|l|l|}
\hline
\textbf{} & \textbf{\indirecth \cite{plets2019efficient}} & \textbf{\name} & \textbf{Improvement (\%)} \\ \hline
Mean error (cm)  & 40.01    & 23.19 & 42.05\% \\ \hline
Median error (cm) & 41.02  & 23.62 & 42.41\% \\ \hline
Max. error (cm)  & 68.98 & 42.52 & 38.35\% \\ \hline
Std. dev. (cm) & 15.68 & 9.64   & 38.53\% \\ \hline
\end{tabular}
\caption{Improvement of positioning accuracy.}
\label{tab:Pos_accuracy_improvement}
\end{table}

\begin{figure*}[h]
    \subfloat[3D Trajectory]{
    \includegraphics[width=0.46\textwidth]{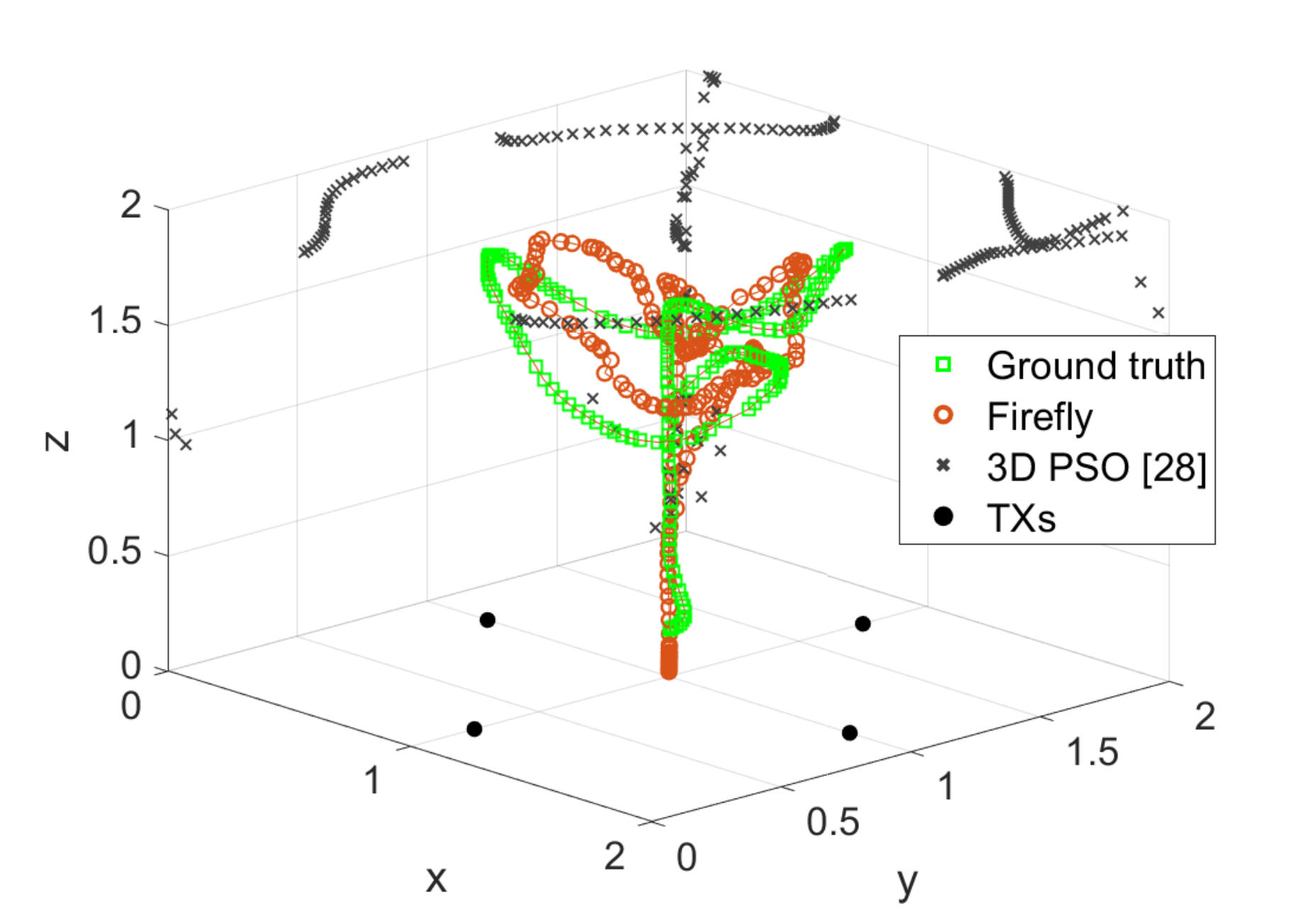}
    \label{fig:route_comparison_3d}
    }
   \subfloat[3D Trajectory (Top View)]{
    \includegraphics[width=0.46\textwidth]{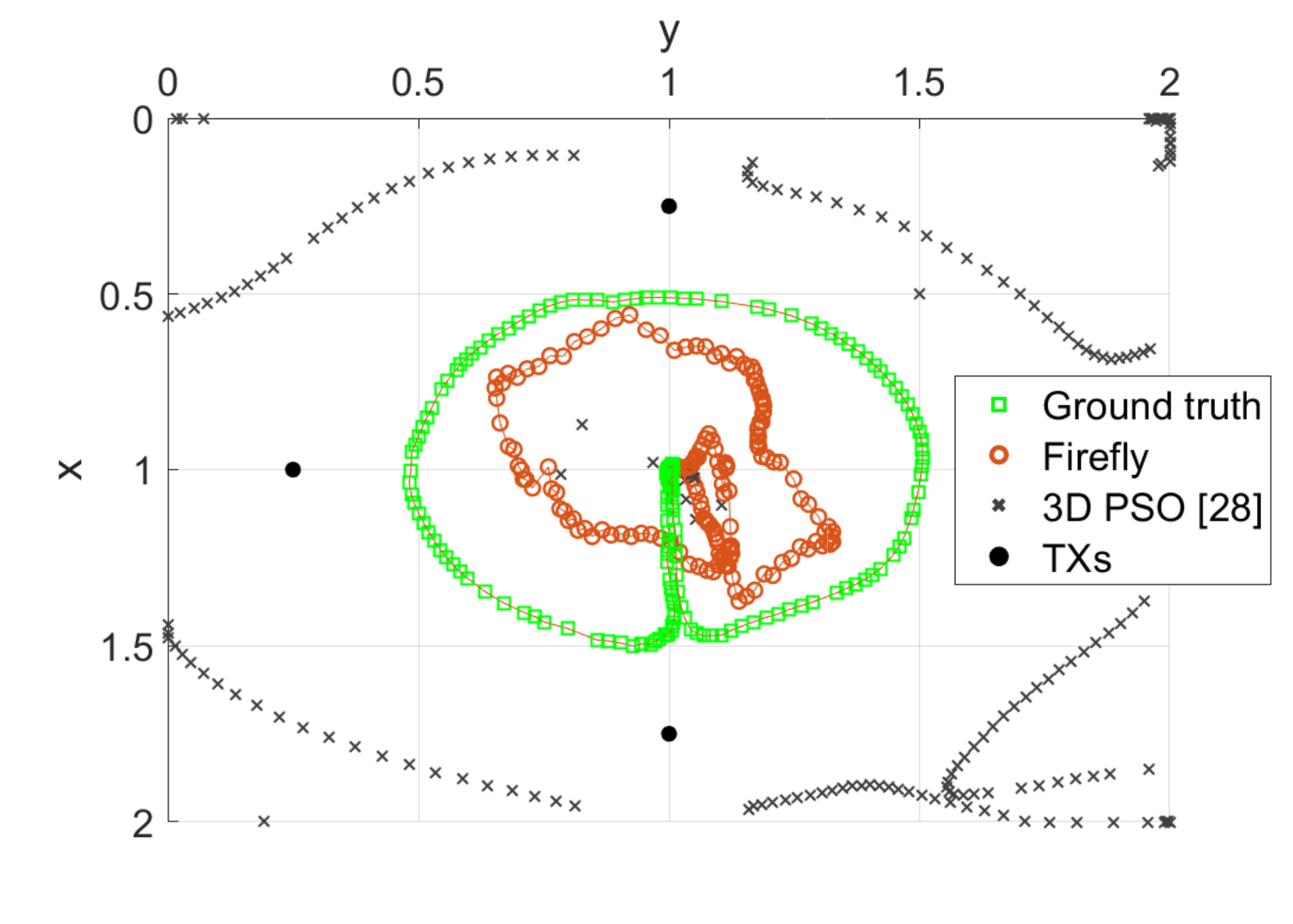}
    \label{fig:route_comparison_3d_top}
    }
    \caption{\name vs. \pso \cite{carreno20-metaheuristic} in Test 8.}
    \label{fig:firefly_vs_pso}
\end{figure*}

\underline{\textit{\name Vs. \pso:}} In another test example, shown in \cref{fig:route_comparison_3d}, we look into the estimated trajectory of the \pso method. \pso displays the expected circular motion in the x-y plane, but the altitude estimation is far from the ground truth, and up to the imposed upper bound of the test area (i.e. \SI{2}{\m}). In \cref{fig:route_comparison_3d_top}, the top view of the trajectory for the same test is shown and we see that the amplitude of the trajectory is very wide compared compared to the ground truth. \pso does not perform as well as the other methods and has a mean positioning error of \SI{94.54}{\cm}.

The inaccurate results of \pso can be explained because its model considers parallel transmitters and receiver. In the method, the gain of the system (i.e. ratio of the received and transmitted power) is used to estimate the $x,y,z$ coordinates that most accurately describe the channel loss (see \cref{eq_chloss}). However, the channel loss cannot be obtained from the power received directly because the distance is also affected by the height of the receiver. When the assumptions of the model are broken, small angle variations can have a significant effect in the distance calculation as the height is not known. Thus, the 3D position cannot be determined precisely.

%Since no height estimation is provided, when the assumption of the model that the transmitters and receiver are parallel is broken, and the model cannot accurately calculate distance. 
%the state space exploration is not constrained to a 2D solution and small angle variation can have a significant effect in the result.

In our test, we show that 3D RSS methods are not effective strategies to determine the position of UAVs. Although they have provided favorable results when tested under controlled conditions, when tilting and movement are introduced, the position accuracy is severely downgraded. \\

\begin{comment}
 \begin{figure*}[h]
 	\begin{minipage}[t]{0.45\linewidth}
         \centering
         \includegraphics[width=0.85\textwidth]{figures/route_comparison_2d_final.pdf}
         \caption{3D Trajectory of Firefly and \indirecth \cite{plets2019efficient} in Test 5.}
         \label{fig:route_comparison_2d}
 	\end{minipage}
 	\hspace{0.5cm}
 	\begin{minipage}[t]{0.45\linewidth}
         \centering
         \includegraphics[width=0.85\textwidth]{figures/route_comparison_2d_final_top.pdf}
         \caption{Trajectory of Firefly and \indirecth \cite{plets2019efficient} in Test 5 (top view).}
         \label{fig:route_comparison_2d_top}
 	\end{minipage}
 	\hfill
 	\vfill
	\begin{minipage}[t]{0.45\linewidth}
 	   \centering
         \includegraphics[width=0.85\textwidth]{figures/route_comparison_3d_final.pdf}
         \caption{3D trajectory of Firefly and \pso \cite{carreno20-metaheuristic} in Test 8.}
         \label{fig:route_comparison_3d}
 	\end{minipage}
 	\hspace{0.5cm}
 	\begin{minipage}[t]{0.45\linewidth}
 	        \centering
         \includegraphics[width=0.85\textwidth]{figures/route_comparison_3d_final_top.pdf}
         \caption{Trajectory of Firefly and \pso \cite{carreno20-metaheuristic} in Test 8 (top view).}
         \label{fig:route_comparison_3d_top}
 	\end{minipage}
 \end{figure*}
 \vspace{0.2cm}
\end{comment}

\subsubsection{Algorithmic complexity}

The \pso method considers multiple particles (candidate solutions) that are evaluated at each iteration. This results in multiple function evaluations and a quadratic complexity $O(n^2)$ without taking into account additional modifications of the method. In~\cite{cai2017-vlp-pso} and~\cite{carreno20-metaheuristic}, 20 iterations of the method are allowed and a particle group size of 200 is considered in the latter study which results in up to 4000 function evaluations. 

In the \indirecth method, the height is computed by iteratively evaluating a set of candidate heights in the range of interest. To evaluate each candidate solution, a trilateration step is performed for each height candidate. Considering the proposed resolution of \SI{1}{\mm} and the height of our setup (\SI{2}{\m}), this would amount to 2000 evaluations. The time complexity of this iterative approach is $O(n)$ if we consider that the computation time of each iteration (consisting of a trilateration step) is constant. A fast search optimization of the algorithm is also proposed in \cite{plets2019efficient} to narrow the search interval. This reduces the computation time by 90\% and the time complexity improves to $O(\log(n))$.

In \name, we implement the complementary filter proposed in \cite{higgins1974-complementary_vs_KF} to estimate the height, which consists of two discrete equations that can be computed in one step respectively. Then, we execute a 2D trilateration method twice. Considering the computation time of the trilateration steps constant (as we do for the \indirecth method), the time complexity of the previous steps is $(O(1))$. When we apply the long term drift correction for the barometer, we also use the \indirecth method to remove zero drift. Therefore, our algorithm has a time complexity of $O(\log(n))$ considering an implementation of the fast search algorithm of \indirecth. However, in our empirical evaluations, we find that the barometer correction can be executed less frequently without a considerable impact on the positioning accuracy. This results in a time complexity of $O(\log(n/k))$, where $k$ indicates that the barometer correction is executed once for every $k$ iterations of our method. When $k=10$, we find that the mean error of \name only increases by \SI{1.6}{\cm}.

\section{Conclusion} %Yanqiu
%The conclusion goes here. novel insights obtained from this research

% general conclusion
This work illustrates, for the first time, that VLC can be used for accurate 3D positioning for drones with six degrees of freedom in a realistic experimental setup. It paves the way for using standard light bulbs to provide navigation services for drones. %This will overcome the limitations of using GPS indoors and in urban canyons, the interference of using RF, as well as privacy concerns and high power consumption of using cameras.
%RQ1: What are the limitation of the SoA?
% answer RQ2: what method should be used
A novel localization method, Firefly, is proposed by building upon the line of research that decomposes a 3D positioning problem into 2D+H. Compared to SoA studies, it removes the need of imposing complex requirements on the transmitter and receiver designs to achieve an accurate positioning result. In addition, it overcomes the limitation of the SoA RSS methods, which are restricted to parallel receiver and transmitters without considering tilting. % and shows detrimental accuracy when tilt is introduced. 
By utilising on-board sensors on the drone, we accurately estimate height and account for tilting of the receiver. The algorithm is also lightweight and can be implemented in a drone for real-time execution. %This facilitates an easy use of standard bulbs as light sources and an online estimation on drones with limited resources. 
% answer RQ3: what is the performance
% testbed; criteria; 
%The results obtained are favorable compared to other SoA approaches...\\
%Mention that only low power LEDs and lowcost sensor were used...\\
%First experimental results gathered from a real 3D flight test subject to movement and tilt.\\
As demonstrated by the experimental results, Firefly achieves mean position accuracy of \SI{23.19}{\cm} using off-the-shelf LED lights and low-cost sensors. It reduces the localization error compared to other SoA methods by around 42\% under the same experimental setup.

\bibliographystyle{IEEEtran}
% argument is your BibTeX string definitions and bibliography database(s)
%\bibliography{IEEEabrv,../bib/paper}
%
% <OR> manually copy in the resultant .bbl file
% set second argument of \begin to the number of references
% (used to reserve space for the reference number labels box)
\addcontentsline{toc}{chapter}{References}
\bibliography{bare_conf} %%this one
%\begin{thebibliography}{1}

%\bibitem{IEEEhowto:kopka}
%H.~Kopka and P.~W. Daly, \emph{A Guide to \LaTeX}, 3rd~ed.\hskip 1em plus
%  0.5em minus 0.4em\relax Harlow, England: Addison-Wesley, 1999.
%\end{thebibliography}

% that's all folks
\end{document}